\documentclass[11pt]{article}

\usepackage[preprint]{acl}

\usepackage{times}
\usepackage{latexsym}
\usepackage[most]{tcolorbox}

\usepackage[T1]{fontenc}

\usepackage[utf8]{inputenc}

\usepackage{microtype}

\usepackage{inconsolata}

\usepackage{graphicx}

\usepackage{amsmath}      
\usepackage{amssymb}      
\usepackage{amsfonts}     
\usepackage{mathtools}    
\usepackage{amsmath, amssymb, amsthm}

\usepackage{algorithm}    
\usepackage{algpseudocode}

\usepackage{natbib}       

\usepackage{enumitem}     
\usepackage{booktabs}     
\usepackage{xcolor}       
\usepackage{hyperref}     
\usepackage{cleveref}     

\usepackage{multirow}

\algrenewcommand\algorithmicrequire{\textbf{Input:}}
\algrenewcommand\algorithmicensure{\textbf{Output:}}
\algnewcommand\Input{\item[\algorithmicrequire]}
\algnewcommand\Output{\item[\algorithmicensure]}

\newtcolorbox{promptbox}[1]{
  breakable, enhanced,
  colback=white, colframe=black!72,
  colbacktitle=black!72, coltitle=white, fonttitle=\bfseries,
  arc=4mm, boxrule=1pt,
  left=3.5mm, right=3.5mm, top=2.5mm, bottom=2.5mm,
  title={#1}
}

\title{POOL: Propagated Uncertainty Over Lookalikes}

\author{
    Rounak Sharma,
    Ananya B. Sai,
    Soumyabrata Pal \\[0.5em]
    Adobe Research, India \\[0.5em]
    \texttt{\{rounaksharma, ananyasai, soumyabratap\}@adobe.com}
}

\begin{document}
\maketitle
\begin{abstract}
Black-box large language models need confidence scores that can separate likely-correct from likely-incorrect outputs, enabling systems to prioritize human review, route uncertain cases to stronger models, or choose abstention thresholds on development data.
Yet existing confidence estimators face a cost-quality trade-off: verbal confidence is cheap but is often overconfident, while sampling-based uncertainty is more informative but scales linearly with the number of samples per query. We propose \textsc{POOL} (\emph{Propagated Uncertainty Over Lookalikes}),
a cost-efficient framework that addresses this trade-off taking inspiration from group-testing.
\textsc{POOL} clusters query stems with overlaps, evaluates a base estimator on representative medoids, softly propagates confidence scores to nearby queries, and selectively evaluates high-disagreement cases.
We instantiate this framework with \textsc{Hy@}$p$, a hybrid estimator that combines verbal confidence with spectral answer diversity computed from the negative von Neumann entropy of sampled answer embeddings. 
Across six domains from three datasets and five black-box LLMs, \textsc{Hy@}5 achieves higher average AUROC than verbal confidence and \textsc{Vn@}10 sampling while using half as many samples as \textsc{Vn@}10. \textsc{POOL}-\textsc{Hy@}5 retains 93.5--97.9\% of its AUROC while saving 19.3--39.3\% of generations. On paraphrase-dense workloads, generation savings rise to 73-76\%, showing that semantic redundancy can be leveraged to lower confidence-estimation costs.

\end{abstract}

\section{Introduction}

Large language models are increasingly deployed in settings where a wrong answer can be very costly. In such systems, confidence scores accompanying the answers play a key role. They determine when to abstain, when to route a query to a stronger model or human expert, how to allocate limited inference budget, or flag outputs for review~\citep{geifman2017selective,kadavath2022language}.

For black-box models accessed only through an API, confidence estimation must satisfy two constraints at once: (accuracy) it must separate likely correct from likely incorrect answers, and (cost) it must be cheap enough to feasibly run over large benchmarks or production-scale query streams.

Existing black-box signals occupy different positions on this trade-off and largely fall into two families.
(i) \textit{Verbal confidence (VC)} asks the model to self-report a numeric confidence value alongside its answer~\citep{xiong2024can,tian-etal-2023-just}. This is cheap, requiring only one generation per query but often suffers from overconfidence and score saturation. Specifically, prior works find that vanilla verbalized confidence values are concentrated in high-confidence bins, often between 80\% and 100\%, even when accuracy is substantially lower~\citep{xiong2024can}. (ii) \textit{Sampling-based estimators} instead draw multiple stochastic answers and measure their agreement or diversity, either 
via semantic clustering \citep{kuhn2023semantic,farquhar2024detecting}, or through spectral properties of answer embeddings~\citep{nikitin2024kernel}. 
These methods are usually more discriminative, especially when verbal confidence collapses, but their cost scales linearly with the number of samples $p$ per query. This makes the stronger confidence estimation approach much less scalable.

We argue that this trade-off is partly an artifact of treating every query independently. Real query batches are rarely arbitrary sets of unrelated inputs. Benchmarks, tutoring systems, customer-support logs, and repeated user workloads often contain questions that share topics, templates, or near-duplicate query stems / cores.

Classical group testing exploits a related principle: when items are structured, expensive tests can be shared across groups and followed by targeted tests only where needed ~\citep{dorfman1943detection}.  \textcolor{black}{At a high level, classical group testing creates overlapping clusters of items, does tests that are representative of entire cluster and finally, for each item, uses results from all clusters the item belongs to infer}.
We adapt these principles to black-box LLM confidence estimation.

We propose \textsc{POOL} (\emph{Propagated Uncertainty Over Lookalikes}), a group-testing-inspired framework for cost-efficient black-box confidence estimation. 
\textcolor{black}{Given any per-query confidence estimator, \textsc{POOL} first clusters query stems in an embedding space (allowing overlaps between clusters) and then evaluates the estimator on cluster medoids to obtain representative scores of the entire cluster. Now, for each query, we propagate \textit{representative scores} from the clusters the query belongs to using soft similarity-based weights - aggregating the propagated scores allows the algorithm to compute the final query-specific confidence. Finally, we spend a small rescue budget on high-disagreement cases, that is, disagreement between the propagated scores, improving the reliability of estimation wherever required.} Since \textsc{POOL} treats the base confidence estimator as a black box, it can serve as a wrapper over off-the-shelf query-specific estimators such as verbal confidence, sampling-based uncertainty without changing their internals.

We analyse the effects of \textsc{POOL} using multiple base query-specific estimators: verbal confidence (\textsc{Vc}), spectral sample-diversity confidence (\textsc{Vn@}$p$), and a simple hybrid of the two we introduce as \textsc{Hy@}$p$. 

The goal of \textsc{Hy@}$p$ is to test whether two standard black-box signals can be combined into a stronger low-cost query-specific method for \textsc{POOL} - empirically speaking, indeed \textsc{Hy@}5 achieves the best average AUROC among the per-query baselines across five datasets and five black-box LLMs. \textcolor{black}{\textit{To the best of our knowledge, the hybrid query-specific estimator of confidence has not been empirically studied in literature.}}
Our main contribution, \textsc{POOL}, preserves most of this ranking quality while reducing the average number of generations per query. With \textsc{Hy@}5 as the base estimator, \textsc{POOL} retains 93.5--97.9\% of unpooled AUROC while saving 39.3\%, 29.3\%, and 19.3\% of confidence-estimation generations as the rescue budget increases. On paraphrase-augmented workloads, savings rise to 73-76\%, confirming that semantic redundancy in the input batch is a usable resource for cheaper confidence estimation.

\textcolor{black}{To summarize, our main contributions are:
\begin{enumerate}[leftmargin=0pt,
    labelindent=0pt,
    itemindent=! ,
    labelsep=0.3em,
    nosep
]
  \item We propose \textsc{POOL}, a method-agnostic framework that reduces black-box confidence-estimation cost by sharing estimator calls across semantic query neighborhoods. Inspired by the classical group testing framework, we follow a similar set of steps namely (1) create overlapping clusters of queries in an embedding space (2) perform representative tests and obtain confidence estimates for the entire cluster as a whole (3) obtain query-specific confidence estimate by aggregating estimates from clusters the query belongs to. 
  \item POOL, in addition, uses \textit{rescue budget} - for queries where the cluster-specific scores have high disagreement, POOL uses additional tests to reduce variance and improve the estimate.  
  \item We evaluate \textsc{POOL} with multiple base query-specific estimators across six domains from three datasets and five LLMs, demonstrating substantial generation savings while preserving most of the query-specific estimator's AUROC. In essence, \textsc{POOL} gives us excellent tradeoff points between AUROC (how good our confidence estimation is) versus the LLM calls/generation cost.  
\end{enumerate}
}

\section{Related Works}

Our work sits at the intersection of confidence estimation for black-box LLMs and cost-efficient group-testing methods. We review the following lines of related work: verbal confidence elicitation, sampling-based uncertainty estimation, and classical group testing.

\textbf{Verbal and prompt-based confidence.}
A growing body of work studies whether LLMs can reliably assess their own uncertainty without access to internal logits.
\citet{kadavath2022language} demonstrated that large language models exhibit partial self-knowledge: when asked whether a given statement is true, they produce calibrated probabilities that improve with scale. \citet{lin2022teaching} trained models to express uncertainty in natural language, while \citet{tian-etal-2023-just} and \citet{xiong2024can} showed that prompting strategies (e.g., chain-of-thought, top-$k$ elicitation) can substantially improve the calibration of self-reported confidence in RLHF-tuned chat models. A common failure mode is \emph{confidence collapse}: instruction-tuned models often report uniformly high confidence, limiting the discriminative power of verbal confidence alone~\citep{xiong2024can}.

\textbf{Sampling-based uncertainty estimation.}
Drawing multiple stochastic generations and measuring their agreement is a natural analogue of deep ensembles~\citep{lakshminarayanan2017simple}, adapted to the setting where one samples from a single model rather than training multiple networks. \citet{wang2023selfconsistency} used majority-vote self-consistency as both an accuracy booster and an implicit confidence signal. \citet{manakul2023selfcheckgpt} proposed \emph{SelfCheckGPT}, which detects hallucinations by measuring inter-sample consistency in a fully black-box setting without reference documents. \citet{kuhn2023semantic} introduced \emph{semantic entropy}, which clusters generations by meaning and computes entropy over cluster probabilities, removing sensitivity to surface-level variation; \citet{farquhar2024detecting} extended this to detect hallucinations in long-form generation. \citet{duan2024shifting} observed that not all tokens contribute equally to semantic meaning and proposed re-weighting token-level uncertainty by relevance. Most closely related to our \textsc{Vn}@$p$ baseline, \citet{nikitin2024kernel} proposed \emph{kernel language entropy}, which computes the Von Neumann entropy of a Gram matrix over answer embeddings, providing a continuous, clustering-free measure of semantic dispersion. Our work builds on their estimator but contributes an orthogonal dimension: we show that hybridizing it with verbal confidence recovers signal when either channel degrades, and that group testing can dramatically reduce the number of generations required at the benchmark level.

\textbf{Group testing and pooled evaluation.}
Classical group testing, introduced by \citet{dorfman1943detection} for efficient screening of blood samples, identifies defective items by testing pools rather than individuals. The theory has been extensively developed~\citep{aldridge2019group} and applied in domains from genomics to communication. To our knowledge, we are the first to apply a group-testing reduction to confidence estimation for LLMs: by clustering semantically similar questions and evaluating only cluster representatives, we convert the per-query cost of any confidence estimator into a per-cluster cost, with the compression ratio determined by the structure of the benchmark.

\section{Method}
\label{sec:method}

We study black-box confidence estimation: given a query $x_i$ from a
batch $\mathcal{X}=\{x_1,\dots,x_N\}$, produce a scalar score
$\hat s_i \in \mathbb{R}$ that is high when the model's answer is
likely correct and low otherwise. We first review five per-query
estimators (\S\ref{sec:base}) that serve as the \emph{base method}
$M$, then introduce \textsc{POOL}, which treats $M$ as a black box
and amortizes its cost across semantically related queries in two
tiers (\S\ref{sec:tier1}--\ref{sec:cost}). All query stems are
embedded by a fixed encoder $\phi(\cdot)$ to unit-norm vectors
$q_i$; similarities are cosine.
We study confidence estimation as a ranking problem, that is, the goal is to assign scores that separate likely-correct from likely-incorrect outputs, not to produce calibrated probabilities.

\subsection{Per-Query Base Estimators}
\label{sec:base}

Let $M : \mathcal{X} \to \mathbb{R}$ be a per-query confidence
estimator with cost $g_M$ generations per call. These estimators are designed to calculate the confidence for each query separately. We consider five
instantiations of such base queries namely Verbal Confidence (\textsc{Vc}), Sample-diversity confidence (\textsc{Vn@}$p$), Hybrid confidence (\textsc{Hy@}$p$), Majority agreement (\textsc{Maj@}$p$), Semantic entropy (\textsc{SE@}$p$). The details of each of the estimators could be found in Appendix \ref{sec:base-estimator}. The results for comparison of base estimators \textsc{Vc}, \textsc{Vn@}$10$, \textsc{Hy@}$5$ with their POOL versions are included in \S\ref{sec:main-results} while the comparison results for the base estimators can be found in Tables \ref{tab:auroc_appendix_vn5_hy10}, \ref{tab:savings_appendix_universal}, \ref{tab:auroc_maj}, \ref{tab:auroc_se}, \ref{tab:auroc_baselines_only} of Appendix \ref{sec:add_res}.

The remainder treats $M$ as a black box with cost $g_M$; any of the
above or any future per-query estimator plugs in unchanged.

\subsection{Tier-1: Representative Selection and Soft Propagation}
\label{sec:tier1}

\textbf{Representative selection.}
We pick $K$ medoid indices
$\mathcal{M}=\{m_1,\dots,m_K\}\subseteq\{1,\dots,N\}$ that cover the
batch in stem-embedding space, by agglomerative clustering with
average linkage and cosine cutoff $\tau\in(0,1)$. For each cluster
$C_k$,
\begin{equation}
  m_k = \arg\max_{i\in C_k}\,
        \tfrac{1}{|C_k|}\sum_{j\in C_k}\langle q_i,q_j\rangle.
\label{eq:medoid}
\end{equation}
$\tau$ controls the compression ratio $c\triangleq K/N\in(0,1]$;
smaller $\tau$ yields fewer, larger clusters. Cluster labels are
discarded after this step --- only the set $\mathcal{M}$ is reused.

\textbf{Base evaluation on medoids.}
We invoke $M$ once per medoid, $s_k = M(x_{m_k})$, consuming $K g_M$
generations --- the Tier-1 floor cost.

\textbf{Soft attention over medoids.}
For each query $i$, identify the medoids within cosine threshold
$\theta\in(0,1)$:
\begin{equation}
  \mathcal{N}_i = \{\,k : \langle q_i, q_{m_k}\rangle \ge \theta\,\},
\label{eq:neighbors}
\end{equation}
falling back to the single nearest medoid if $\mathcal{N}_i=
\emptyset$ and capping at the top $k_{\max}$ by similarity. With
$u_{ik}\triangleq\langle q_i,q_{m_k}\rangle$ and temperature $T>0$,
\begin{equation}
  w_k(i) = \frac{\exp(u_{ik}/T)}
                {\sum_{k'\in\mathcal{N}_i}\exp(u_{ik'}/T)},
\label{eq:weights}
\end{equation}
and the Tier-1 estimate is
\begin{equation}
  \mu_i = \sum_{k\in\mathcal{N}_i} w_k(i)\, s_k.
\label{eq:soft-estimate}
\end{equation}
Each query draws signal from \emph{every} sufficiently close medoid
rather than a single hard partition: queries near boundaries blend
from both sides.

\textbf{Disagreement signal (free).}
The same weights yield a per-query reliability indicator,
\begin{equation}
  d_i = \sqrt{\textstyle\sum_{k\in\mathcal{N}_i} w_k(i)(s_k-\mu_i)^2},
\label{eq:disagreement}
\end{equation}
the weighted standard deviation of the medoid scores feeding $\mu_i$.
Small $d_i$ means the neighbours agreed and $\mu_i$ is faithful;
large $d_i$ means the soft average hides a conflict. No extra
compute is required, $d_i$ reuses the values already in
Eqs.~\ref{eq:weights}--\ref{eq:soft-estimate}.

\subsection{Tier-2: Disagreement-Triggered Rescue}
\label{sec:tier2}

Given a user-controlled rescue fraction $q\in[0,1\!-\!c]$, the rescue
set is the top-$\lceil qN\rceil$ non-medoid queries by $d_i$:
\begin{equation}
  \mathcal{R} = \mathrm{top}_{\lceil qN\rceil}
                \bigl\{ d_i : i \notin \mathcal{M} \bigr\}.
\label{eq:rescue}
\end{equation}
We evaluate $M$ directly on each rescued query and overwrite its
soft estimate. The final per-query score is
\begin{equation}
  \hat s_i = \begin{cases}
    s_{k(i)}, & i\in\mathcal{M}, \ m_{k(i)}=i, \\[2pt]
    M(x_i),    & i\in\mathcal{R}, \\[2pt]
    \mu_i,     & \text{otherwise}.
  \end{cases}
\label{eq:final}
\end{equation}
Allocating Tier-2 query-by-query is strictly finer than at the
cluster level: a single high-disagreement query inside an
otherwise-confident cluster is rescued individually, while uniformly
confident clusters consume no extra budget regardless of size.

\begin{algorithm*}[t]
\caption{\textsc{POOL}: two-tier wrapper around a base confidence
estimator $M$.}
\label{alg:pool}
\small
\begin{algorithmic}[1]
\Require batch $\{x_i\}_{i=1}^{N}$, stem embeddings $\{q_i\}$, base
         estimator $M$ with cost $g_M$;
         hyperparameters $(\tau, \theta, T, k_{\max}, q)$.
\Ensure  per-query scores $\{\hat s_i\}_{i=1}^{N}$.
\Statex
\Statex \textbf{Tier-1: representative selection and soft propagation}
\State $\{C_1,\dots,C_K\} \gets$ AgglomerativeCluster($\{q_i\}$,
       cutoff $1\!-\!\tau$, average linkage)
       \Comment{$K$ disjoint clusters of similar queries}
\For{$k = 1, \dots, K$}
  \State $m_k \gets \arg\max_{i \in C_k}\frac{1}{|C_k|}
         \sum_{j \in C_k}\langle q_i, q_j\rangle$
         \Comment{medoid: most central member}
  \State $s_k \gets M(x_{m_k})$
         \Comment{$K$ base-method calls; the Tier-1 floor cost}
\EndFor
\For{$i = 1, \dots, N$}
  \State $\mathcal{N}_i \gets
         \{k : \langle q_i, q_{m_k}\rangle \ge \theta\}$,
         capped at top-$k_{\max}$ by similarity
         \Comment{soft neighbour set; may contain $> 1$ medoid}
  \If{$\mathcal{N}_i = \emptyset$}\
       $\mathcal{N}_i \gets \{\arg\max_k \langle q_i, q_{m_k}\rangle\}$
       \Comment{fallback for isolated queries}
  \EndIf
  \State $w_k(i) \gets \mathrm{softmax}_{k \in \mathcal{N}_i}
         \bigl(\langle q_i, q_{m_k}\rangle / T\bigr)$
         \Comment{weights peak on the closest medoid}
  \State $\mu_i \gets \sum_{k \in \mathcal{N}_i} w_k(i)\, s_k$
         \Comment{soft estimate: weighted avg of neighbour scores}
  \State $d_i \gets
         \sqrt{\sum_{k \in \mathcal{N}_i} w_k(i)(s_k - \mu_i)^2}$
         \Comment{disagreement: free reliability indicator}
\EndFor
\Statex
\Statex \textbf{Tier-2: disagreement-triggered rescue}
\State $\mathcal{R} \gets$ top-$\lceil qN\rceil$ non-medoid queries
       ranked by $d_i$
       \Comment{queries whose neighbours disagree most}
\For{$r \in \mathcal{R}$}
  \State evaluate $M(x_r)$
         \Comment{$\lceil qN\rceil$ extra base-method calls}
\EndFor
\Statex
\Statex \textbf{Assemble outputs}
\For{$i = 1, \dots, N$}
  \If{$i = m_k$ for some $k$}\ $\hat s_i \gets s_k$
         \Comment{medoid: already evaluated in Tier-1}
  \ElsIf{$i \in \mathcal{R}$}\ $\hat s_i \gets M(x_i)$
         \Comment{rescued: true score from Tier-2}
  \Else\ \ $\hat s_i \gets \mu_i$
         \Comment{otherwise: soft estimate}
  \EndIf
\EndFor
\State \Return $\{\hat s_i\}_{i=1}^{N}$
\end{algorithmic}
\end{algorithm*}

\subsection{Cost and Hyperparameters}
\label{sec:cost}

The total cost is $K + \lceil qN\rceil$ base-method calls, giving
\begin{equation}
  \mathbb{E}[\text{cost per query}] = (c+q)\, g_M,
  \qquad c \triangleq K/N.
\label{eq:cost}
\end{equation}
At $q=0$ the cost collapses to the Tier-1 floor $c\,g_M$; at
$q=1-c$ every non-medoid query is rescued and the un-pooled base
estimator $g_M$ is recovered. \textsc{POOL} has five
hyperparameters:
\begin{center}\small
\begin{tabular}{@{}l@{\ \ }p{0.74\columnwidth}@{}}
$\tau$      & cluster cutoff; controls $K$ (Tier-1 budget). \\
$\theta$    & soft-attention threshold; defines $\mathcal{N}_i$. \\
$T$         & softmax temperature; peakiness of attention. \\
$k_{\max}$  & per-query attention cap. \\
$q$         & Tier-2 rescue fraction; cost--quality dial. \\
\end{tabular}
\end{center}
$\tau$ (which sets the Tier-1 floor cost via $c=K/N$) and $q$ (the
Tier-2 operating point) are deployment-level cost knobs that trade
AUROC for compute by construction and are not tuned.

\subsection{Algorithm}
\label{sec:algorithm}

Algorithm~\ref{alg:pool} summarizes the full procedure. Clustering and
similarity computations are performed once per batch on embeddings; only the $K + \lceil qN\rceil$ base-method calls incur generation cost.

\section{Experiments}
\label{sec:experiments}

\begin{figure*}[t]
  \centering
  \includegraphics[width=\textwidth]{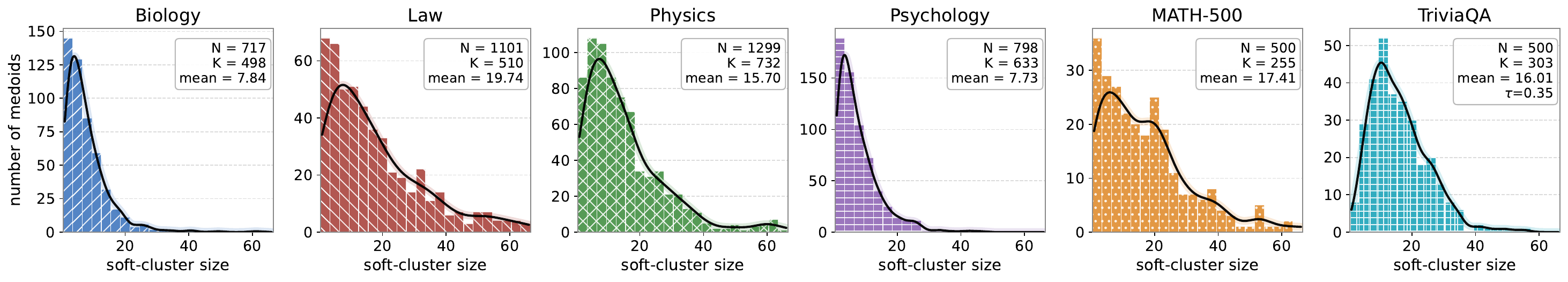}
  \caption{Soft-cluster size distribution (members per medoid under
    POOL's attention rule) for all
    benchmarks. Heavier right tails (Law, Physics,
    MATH-500) reflect denser topical redundancy and translate directly
    into higher \textsc{POOL} compression (Table~\ref{tab:savings_main}).
    Psychology has the lightest tail and correspondingly the smallest
    savings.}
  \label{fig:soft-cluster-hist}
\end{figure*}

\textbf{Datasets.}
We evaluate on six domains across three benchmarks. From \textbf{MMLU-Pro}
\citep{wang2024mmluprorobustchallengingmultitask} we use four
domains --- \textit{biology}, \textit{physics}, \textit{law},
\textit{psychology} --- spanning fact-heavy and reasoning-heavy
multiple-choice. To complement the bounded MCQ format we add
\textbf{MATH-500} \citep{lightman2023lets}, a competition-math
benchmark with free-form numeric answers and \textbf{TriviaQA} \citep{joshi2017triviaqalargescaledistantly} an open-domain trivia benchmark of free-form short-answer questions answered closed-book (no supporting passage).

\textbf{Models.}
We use five black-box LLMs across three families:
\textbf{GPT-4.1-nano}, \textbf{GPT-5-nano}, \textbf{GPT-5-mini},
\textbf{Claude-Haiku-4.5}, \textbf{Llama-3.3-70B-Instruct}.
GPT-5-nano and GPT-5-mini run with reasoning at the \textit{low}
effort setting; the rest are non-reasoning. Stems and answers are
embedded with black-box embedding model \textbf{text-embedding-3-small} (\S\ref{sec:main-results}). We also test POOL with open weight embedding models \textbf{bge-large-en-v1.5}, \textbf{e5-large-v2} (Appendix \ref{sec:add_embed_models}) to validate their generalization across different embedding models. For paraphrase query and adversarial-twin query generation in \S\ref{sec:adv} and \S\ref{sec:paraphrase} we used GPT-5 in reasoning mode.

\textbf{Baselines.}
We evaluate against two categories of baselines. The first compares POOL's
complete two-tier pipeline, end-to-end, against alternative estimators of
comparable cost: \textbf{rand\,$+$\,kNN} (random anchor selection with
$k$-nearest-neighbour propagation), the supervised probes \textbf{P(IK)-LR} and
\textbf{P(IK)-MLP\,(q$+$a)} \citep{kadavath2022language}, and
\textbf{APRICOT} \citep{ulmer-etal-2024-calibrating}. The second category
isolates the rescue mechanism: each method retains POOL's Tier-1 soft
propagation but replaces its Tier-2 rescue rule with an alternative selector,
re-evaluating the queries chosen at random, those nearest a cluster boundary, or
those that inherited the lowest or the highest propagated confidence. We
compare all four against POOL's disagreement-based rescue. The details of the baselines could be found in Appendix \ref{sec:baselines}. The comparison results of POOL with end-to-end baselines are in Table \ref{tab:e2e_baselines} and comparison with rescue strategies is present in Table \ref{tab:rescue_baselines}. 

\textbf{Metric} We evaluate our method using two metrics namely \textbf{AUROC} and \textbf{ECE} commonly used for confidence estimation in literature. The details for the metrics are in Appendix \ref{sec:metric_details}. Detailed results comparing baselines with our proposed framework POOL on AUROC are in Table \ref{tab:auroc_main} and those on ECE are presented in Table \ref{tab:ece_table}.

\textbf{Setup.}
To isolate ranking quality from sampling noise in the underlying
answer, the correctness label for each query is taken from the
\emph{first} sampled answer and reused across all methods. 

\begin{figure}[t]
  \centering
  \includegraphics[width=\columnwidth]{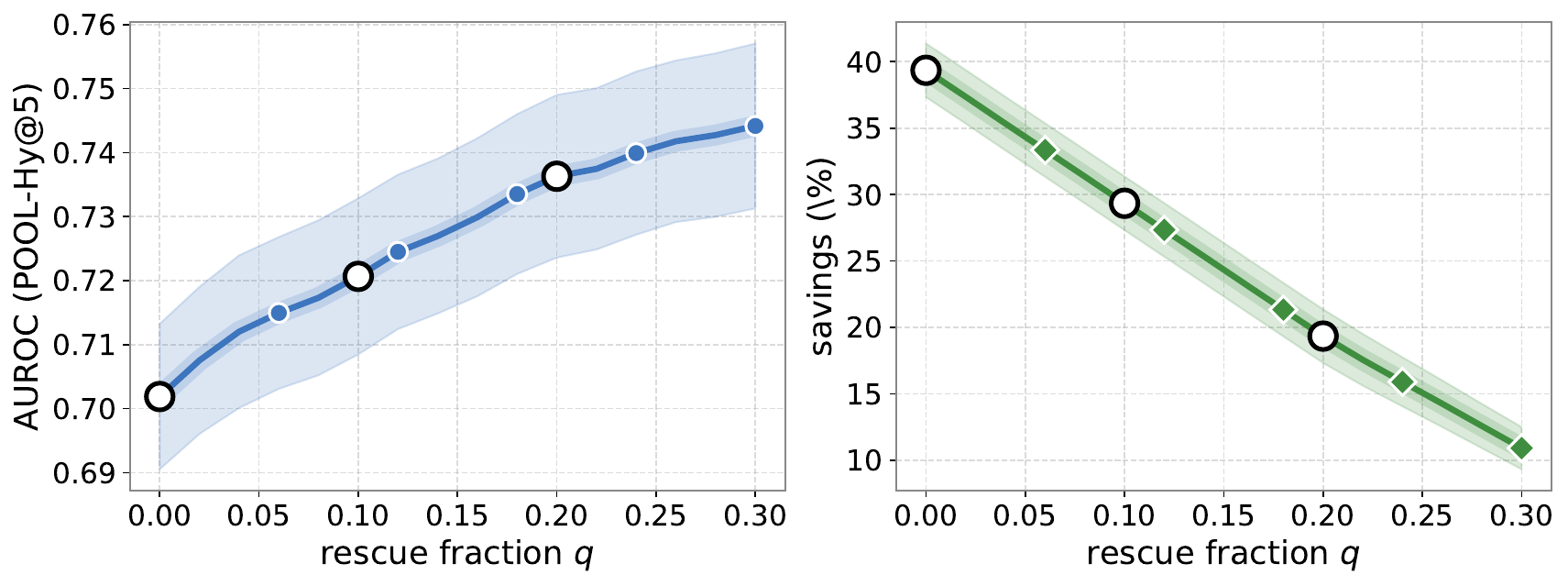}
  \caption{The two faces of the rescue fraction $q$, averaged across
    all 30 (model, dataset) cells. \textbf{Left:} POOL-Hy@5 AUROC
    rises smoothly with $q$. \textbf{Right:} cost savings vs.\ unpooled
    \textsc{Hy@}5 fall linearly with $q$. Open circles mark the three
    operating points $q\in\{0,\,0.10,\,0.20\}$ used in
    Table~\ref{tab:auroc_main}; $q$ is a single, predictable
    cost-quality dial.}
  \label{fig:q-dual}
\end{figure}

\subsection{Main Results}
\label{sec:main-results}

Table~\ref{tab:auroc_main} reports per-cell AUROC for the three base
estimators and six \textsc{POOL} variants across all (model, dataset)
cells. Table~\ref{tab:savings_main} reports the corresponding cost
savings, averaged across the models for each dataset and expressed
relative to each method's own base estimator.

\begin{figure*}[t]
  \centering
  \includegraphics[width=\textwidth]{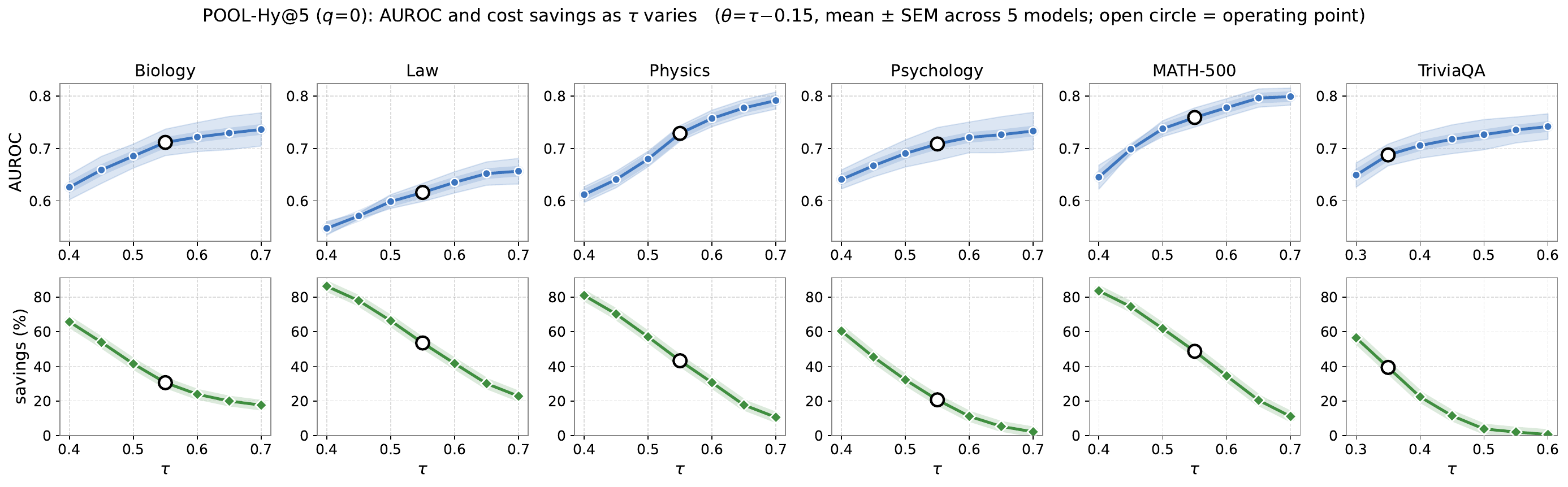}
  \caption{The two faces of the cluster-cutoff knob $\tau$. Top:
    AUROC of POOL-Hy@5 ($q\!=\!0$) rises monotonically with $\tau$.
    Bottom: cost savings vs.\ un-pooled \textsc{Hy@}5 fall monotonically
    with $\tau$. Each line is the mean across all five models with
    shaded SEM. The mirror-image
    behaviour confirms the cost equation $(c+q)\,g_M$: $\tau$ trades
    quality against compute on a smooth, predictable curve.}
  \label{fig:tau-auroc-savings}
\end{figure*}

\begin{table*}[!h]
\centering
\resizebox{\textwidth}{!}{%
\begin{tabular}{l l | ccc | ccc | ccc}
\toprule
& & \multicolumn{3}{c|}{\textbf{Baselines}}
  & \multicolumn{3}{c|}{\textbf{\textsc{POOL}-\textsc{Vn@}10}}
  & \multicolumn{3}{c}{\textbf{\textsc{POOL}-\textsc{Hy@}5}} \\
Model & Dataset
& \textsc{Vc} & \textsc{Vn@}10 & \textsc{Hy@}5
& $q\!=\!0$ & $q\!=\!0.10$ & $q\!=\!0.20$
& $q\!=\!0$ & $q\!=\!0.10$ & $q\!=\!0.20$ \\
\midrule
\multirow{6}{*}{GPT-4.1-nano}
& Biology    & 0.679 & 0.754 & 0.758 & 0.698 & 0.723 & 0.744 & 0.713 & 0.732 & 0.762 \\
& Law        & 0.530 & 0.641 & 0.599 & 0.596 & 0.605 & 0.616 & 0.567 & 0.569 & 0.572 \\
& Physics    & 0.746 & 0.777 & 0.798 & 0.707 & 0.729 & 0.747 & 0.718 & 0.744 & 0.773 \\
& Psychology & 0.630 & 0.744 & 0.716 & 0.730 & 0.743 & 0.753 & 0.700 & 0.702 & 0.726 \\
& MATH-500   & 0.700 & 0.896 & 0.869 & 0.822 & 0.837 & 0.843 & 0.804 & 0.813 & 0.812 \\
& TriviaQA   & 0.686 & 0.761 & 0.748 & 0.672 & 0.707 & 0.721 & 0.668 & 0.695 & 0.723 \\
\midrule
\multirow{6}{*}{GPT-5-mini}
& Biology    & 0.745 & 0.727 & 0.788 & 0.661 & 0.678 & 0.716 & 0.707 & 0.747 & 0.754 \\
& Law        & 0.696 & 0.725 & 0.746 & 0.666 & 0.668 & 0.691 & 0.671 & 0.692 & 0.701 \\
& Physics    & 0.770 & 0.781 & 0.803 & 0.701 & 0.707 & 0.715 & 0.721 & 0.743 & 0.773 \\
& Psychology & 0.773 & 0.728 & 0.796 & 0.705 & 0.713 & 0.734 & 0.752 & 0.783 & 0.792 \\
& MATH-500   & 0.550 & 0.794 & 0.725 & 0.722 & 0.734 & 0.760 & 0.696 & 0.710 & 0.720 \\
& TriviaQA   & 0.683 & 0.723 & 0.715 & 0.676 & 0.679 & 0.691 & 0.688 & 0.712 & 0.710 \\
\midrule
\multirow{6}{*}{GPT-5-nano}
& Biology    & 0.773 & 0.785 & 0.819 & 0.745 & 0.764 & 0.787 & 0.751 & 0.780 & 0.816 \\
& Law        & 0.633 & 0.726 & 0.713 & 0.647 & 0.677 & 0.687 & 0.632 & 0.661 & 0.686 \\
& Physics    & 0.834 & 0.792 & 0.860 & 0.729 & 0.730 & 0.743 & 0.784 & 0.803 & 0.821 \\
& Psychology & 0.730 & 0.742 & 0.774 & 0.731 & 0.747 & 0.770 & 0.752 & 0.766 & 0.784 \\
& MATH-500   & 0.663 & 0.805 & 0.811 & 0.755 & 0.792 & 0.780 & 0.744 & 0.773 & 0.789 \\
& TriviaQA   & 0.754 & 0.796 & 0.810 & 0.742 & 0.771 & 0.767 & 0.751 & 0.780 & 0.783 \\
\midrule
\multirow{6}{*}{Claude-Haiku-4.5}
& Biology    & 0.783 & 0.727 & 0.800 & 0.699 & 0.716 & 0.727 & 0.766 & 0.793 & 0.813 \\
& Law        & 0.658 & 0.652 & 0.672 & 0.601 & 0.623 & 0.632 & 0.608 & 0.629 & 0.630 \\
& Physics    & 0.815 & 0.671 & 0.801 & 0.621 & 0.634 & 0.644 & 0.725 & 0.751 & 0.776 \\
& Psychology & 0.766 & 0.714 & 0.790 & 0.701 & 0.702 & 0.715 & 0.749 & 0.777 & 0.792 \\
& MATH-500   & 0.803 & 0.834 & 0.829 & 0.783 & 0.765 & 0.763 & 0.780 & 0.783 & 0.797 \\
& TriviaQA   & 0.741 & 0.759 & 0.773 & 0.675 & 0.699 & 0.709 & 0.704 & 0.731 & 0.749 \\
\midrule
\multirow{6}{*}{Llama-3.3-70B}
& Biology    & 0.583 & 0.646 & 0.629 & 0.626 & 0.627 & 0.640 & 0.621 & 0.615 & 0.610 \\
& Law        & 0.551 & 0.608 & 0.613 & 0.564 & 0.575 & 0.578 & 0.602 & 0.609 & 0.615 \\
& Physics    & 0.716 & 0.752 & 0.778 & 0.681 & 0.703 & 0.720 & 0.696 & 0.732 & 0.750 \\
& Psychology & 0.551 & 0.597 & 0.598 & 0.585 & 0.605 & 0.635 & 0.590 & 0.600 & 0.610 \\
& MATH-500   & 0.546 & 0.774 & 0.749 & 0.788 & 0.754 & 0.761 & 0.771 & 0.774 & 0.788 \\
& TriviaQA   & 0.589 & 0.697 & 0.654 & 0.642 & 0.664 & 0.662 & 0.626 & 0.622 & 0.663 \\
\midrule
\multicolumn{2}{l|}{\textbf{Average}}
& 0.689 & 0.737 & 0.752 & 0.689 & 0.706 & 0.718 & 0.702 & 0.721 & 0.736 \\
\bottomrule
\end{tabular}%
}
\caption{AUROC across all (model, dataset) cells for the three
per-query baselines (\textsc{Vc}, \textsc{Vn@}10, \textsc{Hy@}5) and
six \textsc{POOL} variants (two base methods $\times$ three rescue
budgets). \textsc{POOL} configuration $(\tau,\theta,T,k_{\max})=
(0.55,\,0.40,\,0.1,\,10)$ is fixed across all cells, \emph{except}
\textbf{TriviaQA}, whose dispersed open-domain stems use $\tau\!=\!0.35$
($\theta\!=\!0.20$) to reach a comparable ${\sim}3$ gen/q operating point
($\tau$ is the cost dial, §3.4); only the rescue fraction $q$ varies.
GPT-5-nano and GPT-5-mini use reasoning effort \textit{low}; the
remaining backbones are non-reasoning models.}
\label{tab:auroc_main}
\end{table*}

\begin{table}[t]\centering\small

\label{tab:ece}
\begin{tabular}{l|cc}
\toprule
Estimator & AUROC & ECE \\
\midrule
\textsc{Hy@}5 (calibrated)             & 0.752 & 0.044 \\
\textsc{POOL}-\textsc{Hy@}5 ($q\!=\!0$)    & 0.702 & \textbf{0.037} \\
\textsc{POOL}-\textsc{Hy@}5 ($q\!=\!0.20$) & 0.736 & 0.038 \\
\bottomrule
\end{tabular}
\caption{Calibration of \textsc{Hy@}5 vs.\ \textsc{POOL}-\textsc{Hy@}5, averaged
over all 30 (model, dataset) cells. ECE uses 15 equal-width
bins; scores are Platt-scaled via 5-fold CV and evaluated out-of-fold.}
\label{tab:ece_table}
\end{table}

\begin{table}[t]\centering\small

\begin{tabular}{ll cc}
\toprule
Method & Paradigm & \multicolumn{2}{c}{AUROC} \\
\cmidrule(lr){3-4}
 & & $q\!=\!0$ & $q\!=\!0.2$ \\
\midrule
P(IK)-LR          & supervised & \multicolumn{2}{c}{0.618} \\
P(IK)-MLP (q$+$a) & supervised & \multicolumn{2}{c}{0.586} \\
APRICOT           & supervised & \multicolumn{2}{c}{0.576} \\
\midrule
rand\,$+$\,kNN    & zero-label & 0.685 & 0.713 \\

\textbf{\textsc{POOL}-\textsc{Hy@}5} & zero-label & 0.702 & \textbf{0.736} \\
\midrule
\textsc{Hy@}5 (un-pooled) & ceiling & \multicolumn{2}{c}{0.752} \\
\bottomrule
\end{tabular}
\caption{\textbf{End-to-end baselines} vs.\ \textsc{POOL}, avg AUROC over 30 cells.
Supervised probes train on labels (no sampling). Zero-label propagation methods
are shown budget-matched at $q\!=\!0$ and $q\!=\!0.2$; the kNN baselines use a
random rescue (no disagreement signal), \textsc{POOL} uses its disagreement rescue.}
\label{tab:e2e_baselines}
\end{table}

\begin{table}[t]\centering\small

\begin{tabular}{l cc}
\toprule
Rescue strategy & AUROC & \% oracle \\
\midrule
Tier-1 only ($q\!=\!0$) & 0.702 & \phantom{0}0\% \\
\midrule
Tier1 + random                & 0.724 & 52\% \\
Tier1 + boundary              & 0.726 & 57\% \\
Tier1 + high-conf             & 0.725 & 55\% \\
Tier1 + low-conf              & 0.728 & 61\% \\
\textbf{Tier1 + disagreement (\textsc{POOL})} & \textbf{0.736} & \textbf{80\%} \\
\midrule
Tier1 + oracle (upper bound)  & \textit{0.745} & \textit{100\%} \\
\bottomrule
\end{tabular}
\caption{\textbf{Rescue-selection strategies} at fixed Tier-1
base $=$ \textsc{Hy@}5 and budget $q\!=\!0.20$, avg over 30
cells. \emph{\% oracle} is the fraction of the oracle rescue's AUROC gain over
Tier-1 that each rule recovers. \textsc{POOL}'s disagreement signal captures the
most.}
\label{tab:rescue_baselines}
\end{table}

\begin{table}[t]
\centering
\small
\setlength{\tabcolsep}{6pt}
\renewcommand{\arraystretch}{1.1}
\begin{tabular}{l | ccc}
\toprule
Dataset & $q\!=\!0$ & $q\!=\!0.10$ & $q\!=\!0.20$ \\
\midrule
Biology     & 30.6\% & 20.6\% & 10.6\% \\
Law         & 53.5\% & 43.5\% & 33.5\% \\
Physics     & 43.2\% & 33.2\% & 23.2\% \\
Psychology  & 20.6\% & 10.6\% &  0.6\% \\
MATH-500    & 48.7\% & 38.7\% & 28.7\% \\
TriviaQA    & 39.4 \% & 29.4 \% & 19.4 \% \\
\midrule
\textbf{Average} & \textbf{39.3\%} & \textbf{29.3\%} & \textbf{19.3\%} \\
\bottomrule
\end{tabular}
\caption{Cost savings of \textsc{POOL} relative to its base
estimator, per dataset, averaged across all models. Each cell reports
$1 - \text{gen/q}_{\text{POOL}} / \text{gen/q}_{\text{base}}$.
Savings are identical for \textsc{POOL}-\textsc{Vn@}10 and
\textsc{POOL}-\textsc{Hy@}5 at any fixed $q$.}
\label{tab:savings_main}
\end{table}

\textbf{Key observations.}
The per-query baselines in Table \ref{tab:auroc_main} show a stable ranking: \textsc{Vc} is weakest
($0.689$ avg AUROC), \textsc{Vn@}10 a clear step above ($0.737$), and
\textsc{Hy@}5 the strongest ($0.752$). The value of \textsc{Hy@}5 is
\emph{efficiency}: with only five samples it matches or beats \textsc{Vn@}10
(ten samples) in $19$ of $30$ cells (mean $+0.014$ AUROC; one-sided
$p\!=\!0.07$).

\textsc{POOL} preserves this ranking while sliding cost along the rescue dial
$q$ (see Section \ref{sec:tier2}). Averaged over all cells, \textsc{POOL}-\textsc{Hy@}5 retains
$93.4\%/95.9\%/97.9\%$ of unpooled \textsc{Hy@}5 AUROC at $39.3\%/29.3\%/19.3\%$
savings ($q\!=\!0/0.10/0.20$), and \textsc{POOL}-\textsc{Vn@}10 traces a parallel
curve ($93.5\%/95.8\%/97.4\%$). The results above vary only the rescue fraction $q$, at a fixed cutoff $\tau$.
In fact $\tau$ (see \S \ref{sec:tier1}) is a second cost--quality knob: it decides how many queries
\textsc{POOL} evaluates directly, so a higher $\tau$ gives higher AUROC but
smaller savings, and a lower $\tau$ gives larger savings but lower AUROC
(see Fig.~\ref{fig:tau-auroc-savings} for effect of $\tau$). We hold $\tau$ fixed per dataset and expose
$q$ as the single user-facing dial.

Per-dataset savings (Table~\ref{tab:savings_main}) track the compression ratio
$K/N$ (see \S \ref{sec:tier1}), which is dataset-dependent but model-independent and set from the
unlabeled stems alone. \textsc{POOL}'s savings scale with stem redundancy. The dense MMLU-Pro/MATH-500
splits give $20$--$54\%$ savings at $\tau\!=\!0.55$ (most on Law/MATH-500, least
on Psychology); TriviaQA's dispersed stems need a looser $\tau\!=\!0.35$ to reach
a comparable $39\%$. In all cases $\tau$ is set from the unlabeled stems: since the compression
$c\!=\!K/N$ depends only on stem geometry and not on correctness, choosing $\tau$
picks a compute budget rather than optimizing the ranking metric. Because $K/N$ is fixed by the stem
clustering, savings are identical for both \textsc{POOL-Vn@10} and \textsc{POOL-Hy@5} variants at any $q$: the
base estimator sets the AUROC ceiling and \textsc{POOL} slides it horizontally by
an amount determined entirely by workload structure.

\subsection{Adversarial Stress Test}
\label{sec:adv}

\textbf{Setup.} To probe POOL under its hardest case, we construct adversarial \emph{twins}:
questions edited by a single word or phrase so the stem stays nearly identical
but the correct answer flips
(GPT-5 generated and answer verified). We add \textbf{946 twins} in total, $445$ for Biology and $501$ for Psychology and run GPT-5-nano on a mixed batch
of the originals plus their twins. A twin nearly duplicates its original's stem,
so it co-clusters with the (confident) original and inherits its propagated
score, a deliberate worst case for stem-based propagation. We report
twin-subset AUROC with cost savings vs.\ unpooled \textsc{Hy@}5. (Table~\ref{tab:adv})

\textbf{Results.} POOL exposes two knobs that both counter this case: a higher
cutoff $\tau$ separates twins from their originals, and the rescue budget $q$
re-evaluates flagged twins directly. Table~\ref{tab:adv} shows the effect is
largely recoverable: at $(\tau,q)\!=\!(0.75,0.20)$ twin AUROC reaches $0.757$
(Biology) and $0.786$ (Psychology) $94$--$96\%$ of the unpooled base, while
still saving $10$--$22\%$ of generations. Robustness to adversarial near-duplicates
is thus controllable through the same deployment dials.

\begin{table}[t]\centering\small
\setlength{\tabcolsep}{6pt}

\begin{tabular}{ll cc cc}
\toprule
& & \multicolumn{2}{c}{Biology} & \multicolumn{2}{c}{Psychology} \\
\cmidrule(lr){3-4}\cmidrule(lr){5-6}
$\tau$ & $q$ & AUROC & sav. & AUROC & sav. \\
\midrule
\multicolumn{2}{l}{Base \textsc{Hy@}5} & 0.808 & -- & 0.819 & -- \\
\midrule
0.55 & 0.00 & 0.669 & 54\% & 0.636 & 47\% \\
0.55 & 0.20 & 0.714 & 34\% & 0.687 & 27\% \\
0.75 & 0.00 & 0.705 & 42\% & 0.674 & 30\% \\
0.75 & 0.20 & \textbf{0.757} & 22\% & \textbf{0.786} & 10\% \\
\bottomrule
\end{tabular}
\caption{\textbf{Adversarial twins (GPT-5-nano).} Twin-subset AUROC and cost
savings of \textsc{POOL}-\textsc{Hy@}5. Raising the cutoff $\tau$ and rescue
budget $q$ recovers most of the base AUROC while retaining savings.}
\label{tab:adv}
\end{table}

\subsection{Test on Paraphrase-Augmented Datasets}
\label{sec:paraphrase}

\textbf{Setup.} Many production workloads (FAQs, customer support, tutoring)
contain semantically near-duplicate questions that the standard MMLU-Pro splits
lack. We build paraphrase-augmented Biology and Psychology splits by asking an LLM
for two paraphrases of each stem, with options and gold answer held fixed;
rephrases whose intended answer changes under a re-ask are discarded. This triples
the density of semantically similar queries at the same nominal difficulty. We run
GPT-5-nano and report AUROC with cost savings vs.\ the unpooled base.

\textbf{Results.} Where the adversarial twins are POOL's worst case, genuine
paraphrases are its best case: the near-duplicates are truly answer-preserving, so
propagation stays accurate and compression is high. At $q\!=\!0$,
\textsc{POOL}-\textsc{Hy@}5 saves $73$--$76\%$ of generations while retaining
${\sim}96\%$ of unpooled AUROC ($0.814\!\to\!0.783$ Biology,
$0.753\!\to\!0.726$ Psychology), and \textsc{POOL}-\textsc{Vn@}10 retains
$97$--$99\%$ (Table~\ref{tab:ext}). POOL's compression thus scales with genuine
cluster density when workloads contain real near-duplicates, the
representative-selection layer turns that structure into a near-proportional
reduction in LLM calls, with no change to the underlying estimator.

\section{Conclusion}

We introduced \textsc{POOL}, a group-testing-inspired framework for cost-efficient black-box confidence estimation. Rather than evaluating a
confidence estimator independently on every query, \textsc{POOL} amortizes estimator calls across semantic neighborhoods.
Since POOL treats the base estimator as a black box, it can be applied to any confidence estimator method.

As a supporting instantiation, we evaluated \textsc{Hy@}$p$, a simple
hybrid base estimator that combines verbal confidence with spectral
sample diversity. \textsc{Hy@}5 provides a strong low-cost base method
for \textsc{POOL}, outperforming the individual black-box signals on
average while using fewer samples than \textsc{Vn@}10. Across six domains from three datasets and five LLMs, \textsc{POOL}-\textsc{Hy@}5 retains
$93.5\text{--}97.9\%$ of unpooled \textsc{Hy@}5 AUROC while saving between $19.3\%$ - $39.3\%$ of generations depending on the rescue budget.
We show the savings on generations in paraphrase-dense workloads rise to $73-76\%$
showing that compression scales with workload redundancy.

Open directions include adaptive medoid selection that adjusts
granularity to local embedding density, alternative rescue triggers
(low absolute soft estimate, or a learned selector), and extending
the framework to open-ended generation.

\section*{Limitations}

\textbf{Batch availability and cold-start.}
POOL operates on a \textit{batch} of queries where the full set is available upfront for clustering, medoid selection, and the score propagation to proceed.
This would be different for the online settings where queries arrive one at a time.
More practically, POOL faces a \textbf{cold-start problem}: on a freshly deployed system with no query history, there are no cached medoid scores to propagate from, and one could end up evaluating every query individually resulting in identical costs to the unpooled base estimator.
The framework's savings accrue only once a sufficient query cache has been built.
We recommend a warm-up phase in which the first $N_{\text{warm}}$ queries are evaluated with the unpooled estimator, their stem embeddings and confidence scores stored, and the medoid structure reconstructed incrementally as the cache grows.
Once the cache reaches a stable compression ratio $c = K/N$, POOL can be activated and subsequent queries routed through the propagation layer.
The required $N_{\text{warm}}$ is workload-dependent: denser topical workloads (e.g., customer-support FAQs) converge faster than sparse, open-domain benchmarks.

\textbf{Input-space propagation assumption.}
We note that POOL currently propagates confidence scores based on cosine similarity between query stem embeddings, implicitly assuming that \textit{similar inputs yield similar confidence profiles}.
While Figure~\ref{fig:rep-confidence} provides aggregate empirical support for this assumption in majority scenarios, it may break down in extreme adversarial scenarios. We stress test this and report the results in Table~\ref{tab:adv}. We also note that this limitation is inherent to any similarity-based approach rather than POOL alone. We believe further research in areas such as embeddings and clustering strategies can further address these limitations.

\bibliography{references}

\appendix

\begin{figure}[t]
  \centering
  \includegraphics[width=\columnwidth]{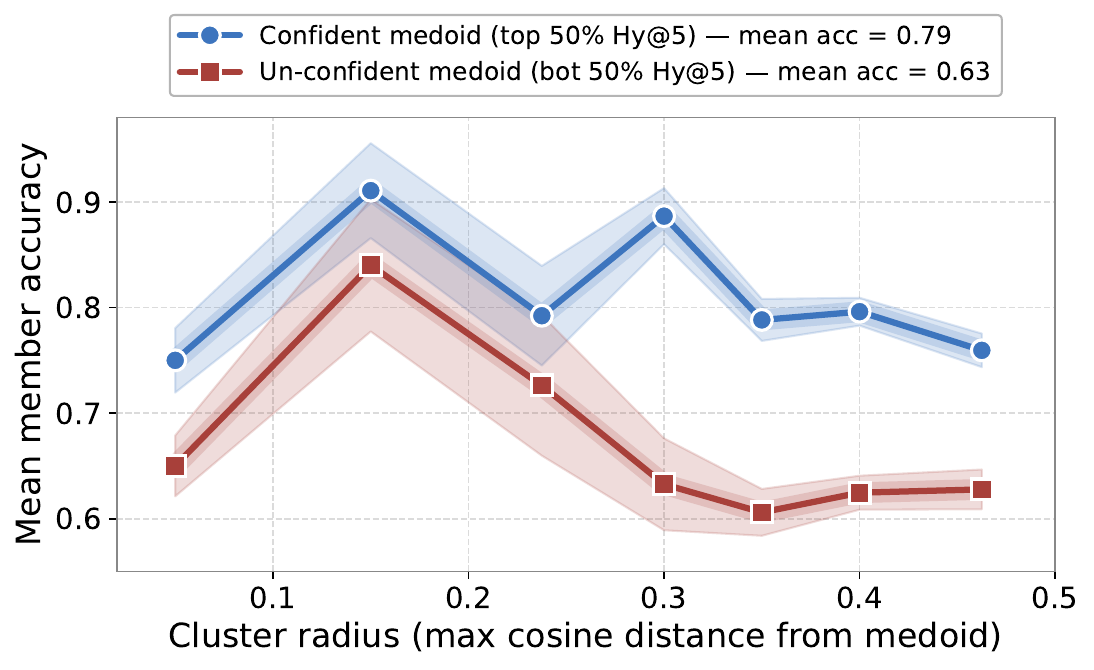}
  \caption{Mean member accuracy of clusters whose medoid lies in the
top vs.\ bottom half of \textsc{Hy@}5 scores, binned by cluster
radius and averaged over all (model, dataset) cells ($\tau=0.55$,
$4{,}442$ multi-member clusters). Members of confident-medoid
clusters are correct $\sim 79\%$ of the time; members of
un-confident-medoid clusters only $\sim 63\%$ --- a $\sim 16$-point
gap that persists across every radius bin and motivates weighting
medoid contributions by their reliability during propagation.}
  \label{fig:rep-confidence}
\end{figure}

\section{Ethics Statement}

This work introduces a method to reduce the cost of confidence estimation for black-box LLMs, with the goal of making selective prediction and abstention mechanisms more accessible.
All experiments use publicly available academic benchmarks (MMLU-Pro, MATH-500, and TriviaQA); no private or personally identifiable data was collected, and no human subjects were involved.

No new models were trained. 
 
POOL's explicit goal is to \emph{reduce} the number of LLM API calls required for confidence estimation.
Table~\ref{tab:cost_breakdown} documents the full per-query cost; at the most expensive operating point ($q\!=\!0.20$), average per-query cost is $\sim$\$2.5$\times 10^{-4}$ using the cheapest backbone.
The clustering and embedding overhead adds less than 3\% to total cost.
We did not train any new models; all computation involved inference calls to existing commercial and open-weight APIs.
We have used AI assistants for writing improvements of the paper.

\section{Additional Results}
\label{sec:add_res}

\begin{table}[t]\centering\small
\setlength{\tabcolsep}{6pt}
\caption{\textbf{Paraphrase-augmented datasets (GPT-5-nano).} AUROC and cost
savings on the $3\times$-density Biology/Psychology splits; \textsc{POOL} at
$(\tau,\theta,T,k_{\max})=(0.55,0.40,0.1,10)$. Savings depend only on the
compression ratio and $q$, so they coincide for both \textsc{POOL} variants.}
\label{tab:ext}
\resizebox{\columnwidth}{!}{%
\begin{tabular}{ll cc cc}
\toprule
& & \multicolumn{2}{c}{Biology-Ext} & \multicolumn{2}{c}{Psychology-Ext} \\
\cmidrule(lr){3-4}\cmidrule(lr){5-6}
Method & $q$ & AUROC & sav. & AUROC & sav. \\
\midrule
\multicolumn{2}{l}{\textsc{Vn@}10 (unpooled)} & 0.773 & -- & 0.736 & -- \\
\multicolumn{2}{l}{\textsc{Hy@}5 (unpooled)}  & 0.814 & -- & 0.753 & -- \\
\midrule
\multirow{3}{*}{\textsc{POOL}-\textsc{Vn@}10}
 & 0.00 & 0.763 & 76\% & 0.716 & 73\% \\
 & 0.10 & 0.767 & 66\% & 0.718 & 63\% \\
 & 0.20 & 0.780 & 56\% & 0.724 & 53\% \\
\midrule
\multirow{3}{*}{\textsc{POOL}-\textsc{Hy@}5}
 & 0.00 & 0.783 & 76\% & 0.726 & 73\% \\
 & 0.10 & 0.793 & 66\% & 0.733 & 63\% \\
 & 0.20 & \textbf{0.810} & 56\% & \textbf{0.730} & 53\% \\
\bottomrule
\end{tabular}%
}
\end{table}

\begin{table*}[t]
\centering
\resizebox{\textwidth}{!}{%
\begin{tabular}{l l | ccc | ccc | ccc}
\toprule
& & \multicolumn{3}{c|}{\textbf{Baselines}}
  & \multicolumn{3}{c|}{\textbf{\textsc{POOL}-\textsc{Vn@}5}}
  & \multicolumn{3}{c}{\textbf{\textsc{POOL}-\textsc{Hy@}10}} \\
Model & Dataset
& \textsc{Vc} & \textsc{Vn@}5 & \textsc{Hy@}10
& $q\!=\!0$ & $q\!=\!0.10$ & $q\!=\!0.20$
& $q\!=\!0$ & $q\!=\!0.10$ & $q\!=\!0.20$ \\
\midrule
\multirow{6}{*}{GPT-4.1-nano}
& Biology     & 0.679 & 0.734 & 0.777 & 0.680 & 0.702 & 0.727 & 0.720 & 0.736 & 0.768 \\
& Law         & 0.530 & 0.618 & 0.611 & 0.576 & 0.587 & 0.601 & 0.577 & 0.585 & 0.591 \\
& Physics     & 0.746 & 0.736 & 0.818 & 0.680 & 0.695 & 0.723 & 0.730 & 0.760 & 0.784 \\
& Psychology  & 0.630 & 0.671 & 0.746 & 0.674 & 0.669 & 0.695 & 0.718 & 0.724 & 0.749 \\
& MATH-500    & 0.700 & 0.858 & 0.882 & 0.804 & 0.807 & 0.828 & 0.816 & 0.832 & 0.829 \\
& TriviaQA    & 0.686 & 0.714 & 0.770 & 0.647 & 0.664 & 0.701 & 0.684 & 0.713 & 0.738 \\
\midrule
\multirow{6}{*}{GPT-5-mini}
& Biology     & 0.745 & 0.695 & 0.775 & 0.650 & 0.664 & 0.695 & 0.707 & 0.752 & 0.758 \\
& Law         & 0.696 & 0.700 & 0.748 & 0.659 & 0.672 & 0.681 & 0.671 & 0.687 & 0.704 \\
& Physics     & 0.770 & 0.691 & 0.832 & 0.664 & 0.649 & 0.673 & 0.740 & 0.763 & 0.793 \\
& Psychology  & 0.773 & 0.645 & 0.809 & 0.649 & 0.648 & 0.668 & 0.765 & 0.790 & 0.803 \\
& MATH-500    & 0.550 & 0.714 & 0.805 & 0.690 & 0.703 & 0.703 & 0.728 & 0.751 & 0.775 \\
& TriviaQA    & 0.683 & 0.681 & 0.742 & 0.673 & 0.668 & 0.695 & 0.696 & 0.720 & 0.723 \\
\midrule
\multirow{6}{*}{GPT-5-nano}
& Biology     & 0.773 & 0.768 & 0.823 & 0.691 & 0.703 & 0.717 & 0.762 & 0.792 & 0.826 \\
& Law         & 0.633 & 0.708 & 0.717 & 0.641 & 0.659 & 0.680 & 0.629 & 0.656 & 0.674 \\
& Physics     & 0.834 & 0.749 & 0.863 & 0.695 & 0.701 & 0.720 & 0.792 & 0.806 & 0.831 \\
& Psychology  & 0.730 & 0.692 & 0.790 & 0.707 & 0.717 & 0.735 & 0.762 & 0.778 & 0.795 \\
& MATH-500    & 0.663 & 0.760 & 0.830 & 0.712 & 0.734 & 0.735 & 0.748 & 0.778 & 0.800 \\
& TriviaQA    & 0.754 & 0.742 & 0.818 & 0.713 & 0.729 & 0.736 & 0.754 & 0.792 & 0.797 \\
\midrule
\multirow{6}{*}{Claude-Haiku-4.5}
& Biology     & 0.783 & 0.690 & 0.807 & 0.679 & 0.689 & 0.717 & 0.771 & 0.790 & 0.812 \\
& Law         & 0.658 & 0.619 & 0.685 & 0.574 & 0.589 & 0.594 & 0.621 & 0.651 & 0.645 \\
& Physics     & 0.815 & 0.688 & 0.791 & 0.638 & 0.660 & 0.664 & 0.720 & 0.741 & 0.762 \\
& Psychology  & 0.766 & 0.693 & 0.795 & 0.684 & 0.694 & 0.698 & 0.757 & 0.777 & 0.790 \\
& MATH-500    & 0.803 & 0.777 & 0.855 & 0.770 & 0.756 & 0.733 & 0.783 & 0.785 & 0.803 \\
& TriviaQA    & 0.741 & 0.691 & 0.798 & 0.640 & 0.641 & 0.658 & 0.719 & 0.741 & 0.773 \\
\midrule
\multirow{6}{*}{Llama-3.3-70B}
& Biology     & 0.583 & 0.615 & 0.645 & 0.601 & 0.595 & 0.613 & 0.629 & 0.630 & 0.623 \\
& Law         & 0.551 & 0.592 & 0.611 & 0.577 & 0.583 & 0.588 & 0.581 & 0.583 & 0.591 \\
& Physics     & 0.716 & 0.715 & 0.798 & 0.650 & 0.668 & 0.684 & 0.714 & 0.748 & 0.769 \\
& Psychology  & 0.551 & 0.571 & 0.609 & 0.554 & 0.559 & 0.579 & 0.597 & 0.615 & 0.626 \\
& MATH-500    & 0.546 & 0.737 & 0.788 & 0.775 & 0.743 & 0.734 & 0.784 & 0.783 & 0.786 \\
& TriviaQA    & 0.589 & 0.622 & 0.711 & 0.604 & 0.612 & 0.636 & 0.643 & 0.659 & 0.680 \\
\midrule
\multicolumn{2}{l|}{\textbf{Average}}
& 0.689 & 0.696 & 0.768 & 0.665 & 0.672 & 0.687 & 0.711 & 0.731 & 0.747 \\
\bottomrule
\end{tabular}%
}
\caption{Appendix AUROC table -- same setup as Table~\ref{tab:auroc_main}
but with \textsc{Vn@}5 (instead of \textsc{Vn@}10) and \textsc{Hy@}10
(instead of \textsc{Hy@}5).  \textsc{POOL} configuration
$(\theta,T,k_{\max})\!=\!(0.40,0.1,10)$ with $\tau\!=\!0.55$ (MMLU-Pro/MATH-500)
and $\tau\!=\!0.35$ (TriviaQA); only $q$ varies.}
\label{tab:auroc_appendix_vn5_hy10}
\end{table*}

\begin{table*}[t]
\centering
\small
\setlength{\tabcolsep}{6pt}
\renewcommand{\arraystretch}{1.1}
\begin{tabular}{l | ccc}
\toprule
Dataset & $q\!=\!0$ & $q\!=\!0.10$ & $q\!=\!0.20$ \\
\midrule
Biology     & 30.6\% & 20.6\% & 10.6\% \\
Law         & 53.5\% & 43.5\% & 33.5\% \\
Physics     & 43.2\% & 33.2\% & 23.2\% \\
Psychology  & 20.6\% & 10.6\% &  0.6\% \\
MATH-500    & 48.7\% & 38.7\% & 28.7\% \\
TriviaQA    & 39.4 \% & 29.4 \% & 19.4 \% \\

\midrule
\textbf{Average} & \textbf{39.3\%} & \textbf{29.3\%} & \textbf{19.3\%} \\
\bottomrule
\end{tabular}
\caption{Cost savings of \textsc{POOL} over its base estimator, per
dataset, averaged across all five models. Savings are identical for
every \textsc{POOL}-$M$ variant ($M\in\{$\textsc{Vn@}5,
\textsc{Vn@}10, \textsc{Hy@}5, \textsc{Hy@}10, \textsc{Maj@}5,
\textsc{Maj@}10, \textsc{SE@}5, \textsc{SE@}10$\}$) at any fixed $q$,
because the per-query cost formula $(c+q)\,g_M$ makes the base-method
factor $g_M$ cancel out in the ratio, the relative reduction
depends only on the clustering compression $c\!=\!K/N$ and the rescue
fraction $q$.}
\label{tab:savings_appendix_universal}
\end{table*}

\begin{table*}[t]
\centering
\resizebox{\textwidth}{!}{%
\begin{tabular}{l l | cc | ccc | ccc}
\toprule
& & \multicolumn{2}{c|}{\textbf{Baselines}}
  & \multicolumn{3}{c|}{\textbf{\textsc{POOL}-\textsc{Maj@}5}}
  & \multicolumn{3}{c}{\textbf{\textsc{POOL}-\textsc{Maj@}10}} \\
Model & Dataset & \textsc{Maj@}5 & \textsc{Maj@}10
& $q\!=\!0$ & $q\!=\!0.10$ & $q\!=\!0.20$
& $q\!=\!0$ & $q\!=\!0.10$ & $q\!=\!0.20$ \\
\midrule
\multirow{6}{*}{GPT-4.1-nano}
& Biology     & 0.742 & 0.772 & 0.710 & 0.730 & 0.759 & 0.724 & 0.748 & 0.772 \\
& Law         & 0.633 & 0.658 & 0.590 & 0.610 & 0.621 & 0.603 & 0.620 & 0.623 \\
& Physics     & 0.761 & 0.802 & 0.703 & 0.723 & 0.728 & 0.724 & 0.746 & 0.765 \\
& Psychology  & 0.674 & 0.735 & 0.665 & 0.664 & 0.688 & 0.723 & 0.737 & 0.744 \\
& MATH-500    & 0.866 & 0.896 & 0.812 & 0.828 & 0.827 & 0.831 & 0.847 & 0.854 \\
& TriviaQA    & 0.717 & 0.758 & 0.656 & 0.666 & 0.701 & 0.682 & 0.701 & 0.726 \\
\midrule
\multirow{6}{*}{GPT-5-mini}
& Biology     & 0.674 & 0.698 & 0.619 & 0.626 & 0.646 & 0.642 & 0.653 & 0.671 \\
& Law         & 0.701 & 0.721 & 0.657 & 0.674 & 0.687 & 0.663 & 0.678 & 0.690 \\
& Physics     & 0.724 & 0.780 & 0.659 & 0.664 & 0.691 & 0.695 & 0.710 & 0.728 \\
& Psychology  & 0.656 & 0.700 & 0.658 & 0.660 & 0.676 & 0.698 & 0.706 & 0.720 \\
& MATH-500    & 0.750 & 0.806 & 0.690 & 0.705 & 0.714 & 0.719 & 0.741 & 0.751 \\
& TriviaQA    & 0.702 & 0.708 & 0.668 & 0.664 & 0.691 & 0.666 & 0.666 & 0.690 \\
\midrule
\multirow{6}{*}{GPT-5-nano}
& Biology     & 0.740 & 0.781 & 0.701 & 0.702 & 0.740 & 0.745 & 0.743 & 0.772 \\
& Law         & 0.709 & 0.735 & 0.654 & 0.691 & 0.696 & 0.664 & 0.677 & 0.697 \\
& Physics     & 0.757 & 0.791 & 0.699 & 0.726 & 0.722 & 0.738 & 0.762 & 0.758 \\
& Psychology  & 0.699 & 0.740 & 0.703 & 0.712 & 0.726 & 0.730 & 0.742 & 0.764 \\
& MATH-500    & 0.748 & 0.797 & 0.719 & 0.737 & 0.745 & 0.747 & 0.786 & 0.788 \\
& TriviaQA    & 0.742 & 0.780 & 0.717 & 0.722 & 0.718 & 0.748 & 0.754 & 0.758 \\
\midrule
\multirow{6}{*}{Claude-Haiku-4.5}
& Biology     & 0.710 & 0.748 & 0.699 & 0.718 & 0.741 & 0.727 & 0.740 & 0.758 \\
& Law         & 0.635 & 0.675 & 0.589 & 0.609 & 0.611 & 0.615 & 0.645 & 0.661 \\
& Physics     & 0.724 & 0.727 & 0.676 & 0.695 & 0.710 & 0.680 & 0.698 & 0.719 \\
& Psychology  & 0.684 & 0.716 & 0.693 & 0.711 & 0.709 & 0.712 & 0.728 & 0.732 \\
& MATH-500    & 0.763 & 0.827 & 0.763 & 0.748 & 0.703 & 0.792 & 0.771 & 0.758 \\
& TriviaQA    & 0.691 & 0.744 & 0.639 & 0.655 & 0.665 & 0.664 & 0.688 & 0.708 \\
\midrule
\multirow{6}{*}{Llama-3.3-70B}
& Biology     & 0.589 & 0.625 & 0.581 & 0.565 & 0.571 & 0.607 & 0.605 & 0.608 \\
& Law         & 0.602 & 0.616 & 0.577 & 0.588 & 0.592 & 0.562 & 0.588 & 0.585 \\
& Physics     & 0.710 & 0.756 & 0.655 & 0.680 & 0.695 & 0.690 & 0.714 & 0.726 \\
& Psychology  & 0.591 & 0.611 & 0.569 & 0.578 & 0.597 & 0.600 & 0.613 & 0.644 \\
& MATH-500    & 0.756 & 0.782 & 0.766 & 0.750 & 0.754 & 0.778 & 0.752 & 0.743 \\
& TriviaQA    & 0.620 & 0.662 & 0.613 & 0.607 & 0.639 & 0.639 & 0.647 & 0.661 \\
\midrule
\multicolumn{2}{l|}{\textbf{Average}}
& 0.702 & 0.738 & 0.670 & 0.680 & 0.692 & 0.693 & 0.707 & 0.719 \\
\bottomrule
\end{tabular}%
}
\caption{Appendix AUROC table for \textsc{Maj@}5 and \textsc{Maj@}10 baselines and their \textsc{POOL} variants. Same setup as Table~\ref{tab:auroc_main}; \textsc{POOL} configuration $(\theta,T,k_{\max})\!=\!(0.40,0.1,10)$ with $\tau\!=\!0.55$ (MMLU-Pro/MATH-500) and $\tau\!=\!0.35$ (TriviaQA); only $q$ varies.}
\label{tab:auroc_maj}
\end{table*}

\begin{table*}[t]
\centering
\resizebox{\textwidth}{!}{%
\begin{tabular}{l l | cc | ccc | ccc}
\toprule
& & \multicolumn{2}{c|}{\textbf{Baselines}}
  & \multicolumn{3}{c|}{\textbf{\textsc{POOL}-\textsc{SE@}5}}
  & \multicolumn{3}{c}{\textbf{\textsc{POOL}-\textsc{SE@}10}} \\
Model & Dataset & \textsc{SE@}5 & \textsc{SE@}10
& $q\!=\!0$ & $q\!=\!0.10$ & $q\!=\!0.20$
& $q\!=\!0$ & $q\!=\!0.10$ & $q\!=\!0.20$ \\
\midrule
\multirow{6}{*}{GPT-4.1-nano}
& Biology     & 0.685 & 0.699 & 0.670 & 0.661 & 0.690 & 0.678 & 0.666 & 0.683 \\
& Law         & 0.619 & 0.638 & 0.583 & 0.587 & 0.603 & 0.599 & 0.604 & 0.604 \\
& Physics     & 0.666 & 0.692 & 0.644 & 0.648 & 0.655 & 0.656 & 0.669 & 0.680 \\
& Psychology  & 0.651 & 0.712 & 0.651 & 0.651 & 0.670 & 0.711 & 0.725 & 0.730 \\
& MATH-500    & 0.839 & 0.863 & 0.797 & 0.793 & 0.818 & 0.818 & 0.821 & 0.830 \\
& TriviaQA    & 0.716 & 0.752 & 0.644 & 0.664 & 0.692 & 0.666 & 0.694 & 0.716 \\
\midrule
\multirow{6}{*}{GPT-5-mini}
& Biology     & 0.618 & 0.650 & 0.562 & 0.574 & 0.591 & 0.583 & 0.605 & 0.615 \\
& Law         & 0.696 & 0.717 & 0.658 & 0.670 & 0.681 & 0.665 & 0.668 & 0.688 \\
& Physics     & 0.657 & 0.685 & 0.623 & 0.601 & 0.629 & 0.635 & 0.641 & 0.647 \\
& Psychology  & 0.653 & 0.694 & 0.656 & 0.665 & 0.676 & 0.694 & 0.703 & 0.711 \\
& MATH-500    & 0.713 & 0.764 & 0.684 & 0.684 & 0.693 & 0.710 & 0.704 & 0.711 \\
& TriviaQA    & 0.674 & 0.681 & 0.648 & 0.637 & 0.677 & 0.657 & 0.656 & 0.687 \\
\midrule
\multirow{6}{*}{GPT-5-nano}
& Biology     & 0.675 & 0.703 & 0.641 & 0.663 & 0.662 & 0.701 & 0.718 & 0.722 \\
& Law         & 0.694 & 0.715 & 0.640 & 0.665 & 0.669 & 0.646 & 0.668 & 0.680 \\
& Physics     & 0.669 & 0.690 & 0.622 & 0.620 & 0.647 & 0.647 & 0.653 & 0.655 \\
& Psychology  & 0.685 & 0.727 & 0.696 & 0.703 & 0.723 & 0.725 & 0.738 & 0.758 \\
& MATH-500    & 0.719 & 0.781 & 0.682 & 0.696 & 0.710 & 0.746 & 0.775 & 0.770 \\
& TriviaQA    & 0.734 & 0.762 & 0.714 & 0.718 & 0.725 & 0.733 & 0.760 & 0.752 \\
\midrule
\multirow{6}{*}{Claude-Haiku-4.5}
& Biology     & 0.663 & 0.682 & 0.666 & 0.685 & 0.702 & 0.670 & 0.669 & 0.690 \\
& Law         & 0.616 & 0.655 & 0.576 & 0.595 & 0.597 & 0.604 & 0.626 & 0.638 \\
& Physics     & 0.660 & 0.647 & 0.632 & 0.640 & 0.648 & 0.612 & 0.617 & 0.632 \\
& Psychology  & 0.670 & 0.698 & 0.678 & 0.682 & 0.695 & 0.694 & 0.695 & 0.711 \\
& MATH-500    & 0.737 & 0.795 & 0.748 & 0.739 & 0.705 & 0.741 & 0.728 & 0.723 \\
& TriviaQA    & 0.662 & 0.708 & 0.627 & 0.627 & 0.635 & 0.638 & 0.657 & 0.655 \\
\midrule
\multirow{6}{*}{Llama-3.3-70B}
& Biology     & 0.573 & 0.585 & 0.592 & 0.573 & 0.578 & 0.607 & 0.588 & 0.590 \\
& Law         & 0.591 & 0.599 & 0.571 & 0.577 & 0.593 & 0.557 & 0.569 & 0.574 \\
& Physics     & 0.635 & 0.667 & 0.606 & 0.629 & 0.647 & 0.631 & 0.651 & 0.658 \\
& Psychology  & 0.580 & 0.600 & 0.555 & 0.564 & 0.580 & 0.584 & 0.601 & 0.626 \\
& MATH-500    & 0.748 & 0.778 & 0.781 & 0.746 & 0.741 & 0.793 & 0.763 & 0.764 \\
& TriviaQA    & 0.612 & 0.645 & 0.590 & 0.586 & 0.616 & 0.643 & 0.657 & 0.659 \\
\midrule
\multicolumn{2}{l|}{\textbf{Average}}
& 0.670 & 0.699 & 0.648 & 0.651 & 0.665 & 0.668 & 0.677 & 0.686 \\
\bottomrule
\end{tabular}%
}
\caption{Appendix AUROC table for \textsc{SE@}5 and \textsc{SE@}10 baselines and their \textsc{POOL} variants. Same setup as Table~\ref{tab:auroc_main}; \textsc{POOL} configuration $(\theta,T,k_{\max})\!=\!(0.40,0.1,10)$ with $\tau\!=\!0.55$ (MMLU-Pro/MATH-500) and $\tau\!=\!0.35$ (TriviaQA); only $q$ varies.}
\label{tab:auroc_se}
\end{table*}

\begin{table*}[t]
\centering
\resizebox{\textwidth}{!}{%
\begin{tabular}{l l | c | cc | cc | cc | cc}
\toprule
Model & Dataset & \textsc{Vc} & \textsc{Vn@}5 & \textsc{Vn@}10 & \textsc{Hy@}5 & \textsc{Hy@}10 & \textsc{Maj@}5 & \textsc{Maj@}10 & \textsc{SE@}5 & \textsc{SE@}10 \\
\midrule
\multirow{6}{*}{GPT-4.1-nano}
& Biology     & 0.679 & 0.734 & 0.754 & 0.758 & 0.777 & 0.742 & 0.772 & 0.685 & 0.699 \\
& Law         & 0.530 & 0.618 & 0.641 & 0.599 & 0.611 & 0.633 & 0.658 & 0.619 & 0.638 \\
& Physics     & 0.746 & 0.736 & 0.777 & 0.798 & 0.818 & 0.761 & 0.802 & 0.666 & 0.692 \\
& Psychology  & 0.630 & 0.671 & 0.744 & 0.716 & 0.746 & 0.674 & 0.735 & 0.651 & 0.712 \\
& MATH-500    & 0.700 & 0.858 & 0.896 & 0.869 & 0.882 & 0.866 & 0.896 & 0.839 & 0.863 \\
& TriviaQA    & 0.686 & 0.714 & 0.761 & 0.748 & 0.770 & 0.717 & 0.758 & 0.716 & 0.752 \\
\midrule
\multirow{6}{*}{GPT-5-mini}
& Biology     & 0.745 & 0.695 & 0.727 & 0.788 & 0.775 & 0.674 & 0.698 & 0.618 & 0.650 \\
& Law         & 0.696 & 0.700 & 0.725 & 0.746 & 0.748 & 0.701 & 0.721 & 0.696 & 0.717 \\
& Physics     & 0.770 & 0.691 & 0.781 & 0.803 & 0.832 & 0.724 & 0.780 & 0.657 & 0.685 \\
& Psychology  & 0.773 & 0.645 & 0.728 & 0.796 & 0.809 & 0.656 & 0.700 & 0.653 & 0.694 \\
& MATH-500    & 0.550 & 0.714 & 0.794 & 0.725 & 0.805 & 0.750 & 0.806 & 0.713 & 0.764 \\
& TriviaQA    & 0.683 & 0.681 & 0.723 & 0.715 & 0.742 & 0.702 & 0.708 & 0.674 & 0.681 \\
\midrule
\multirow{6}{*}{GPT-5-nano}
& Biology     & 0.773 & 0.768 & 0.785 & 0.819 & 0.823 & 0.740 & 0.781 & 0.675 & 0.703 \\
& Law         & 0.633 & 0.708 & 0.726 & 0.713 & 0.717 & 0.709 & 0.735 & 0.694 & 0.715 \\
& Physics     & 0.834 & 0.749 & 0.792 & 0.860 & 0.863 & 0.757 & 0.791 & 0.669 & 0.690 \\
& Psychology  & 0.730 & 0.692 & 0.742 & 0.774 & 0.790 & 0.699 & 0.740 & 0.685 & 0.727 \\
& MATH-500    & 0.663 & 0.760 & 0.805 & 0.811 & 0.830 & 0.748 & 0.797 & 0.719 & 0.781 \\
& TriviaQA    & 0.754 & 0.742 & 0.796 & 0.810 & 0.818 & 0.742 & 0.780 & 0.734 & 0.762 \\
\midrule
\multirow{6}{*}{Claude-Haiku-4.5}
& Biology     & 0.783 & 0.690 & 0.727 & 0.800 & 0.807 & 0.710 & 0.748 & 0.663 & 0.682 \\
& Law         & 0.658 & 0.619 & 0.652 & 0.672 & 0.685 & 0.635 & 0.675 & 0.616 & 0.655 \\
& Physics     & 0.815 & 0.688 & 0.671 & 0.801 & 0.791 & 0.724 & 0.727 & 0.660 & 0.647 \\
& Psychology  & 0.766 & 0.693 & 0.714 & 0.790 & 0.795 & 0.684 & 0.716 & 0.670 & 0.698 \\
& MATH-500    & 0.803 & 0.777 & 0.834 & 0.829 & 0.855 & 0.763 & 0.827 & 0.737 & 0.795 \\
& TriviaQA    & 0.741 & 0.691 & 0.759 & 0.773 & 0.798 & 0.691 & 0.744 & 0.662 & 0.708 \\
\midrule
\multirow{6}{*}{Llama-3.3-70B}
& Biology     & 0.583 & 0.615 & 0.646 & 0.629 & 0.645 & 0.589 & 0.625 & 0.573 & 0.585 \\
& Law         & 0.551 & 0.592 & 0.608 & 0.613 & 0.611 & 0.602 & 0.616 & 0.591 & 0.599 \\
& Physics     & 0.716 & 0.715 & 0.752 & 0.778 & 0.798 & 0.710 & 0.756 & 0.635 & 0.667 \\
& Psychology  & 0.551 & 0.571 & 0.597 & 0.598 & 0.609 & 0.591 & 0.611 & 0.580 & 0.600 \\
& MATH-500    & 0.546 & 0.737 & 0.774 & 0.749 & 0.788 & 0.756 & 0.782 & 0.748 & 0.778 \\
& TriviaQA    & 0.589 & 0.622 & 0.697 & 0.654 & 0.711 & 0.620 & 0.662 & 0.612 & 0.645 \\
\midrule
\multicolumn{2}{l|}{\textbf{Average}}
& 0.689 & 0.696 & 0.737 & 0.752 & 0.768 & 0.702 & 0.738 & 0.670 & 0.699 \\
\bottomrule
\end{tabular}%
}
\caption{All nine per-query (point-based) confidence estimators evaluated across five models and \textbf{six} datasets. \textsc{Vc} = verbal confidence; \textsc{Vn@}$p$ = spectral sample-diversity (negative von Neumann entropy of $p$ answer embeddings); \textsc{Hy@}$p$ = z-scored sum of \textsc{Vc} and \textsc{Vn@}$p$; \textsc{Maj@}$p$ = self-consistency / majority-vote fraction; \textsc{SE@}$p$ = discrete semantic entropy (Shannon entropy over union-find clusters of answer embeddings, $\eta\!=\!0.85$). GPT-5-nano and GPT-5-mini use reasoning effort \textit{low}; the rest are non-reasoning. Each method costs $p$ generations per query (\textsc{Vc} costs $1$).}
\label{tab:auroc_baselines_only}
\end{table*}

\textbf{The hybrid combination choice.} Table~\ref{tab:ablation_hy} ablates the $z$-scored sum
that defines \textsc{Hy@}5. Fusing the two channels lifts AUROC by
$\sim 0.06$ over either alone, confirming that \textsc{Vc} and
\textsc{Vn@}5 carry complementary error modes. Within the space of
linear combinations, performance is essentially flat -- $\alpha
\in [0.25, 0.75]$ all score within $0.008$ of the equal-weight
default ($0.745$--$0.753$), so the choice $\alpha\!=\!0.5$ removes
one hyperparameter without cost. The functional form matters more
than the weighting: replacing the sum with $\max(\cdot,\cdot)$
loses $0.04$ AUROC (a single over-confident channel can promote
wrong answers), while $\min(\cdot,\cdot)$ does nearly as well as
sum. Finally, $z$-scoring is essential: dropping it costs $0.017$
AUROC because the two channels live on incompatible scales and
\textsc{Vn} otherwise dominates the sum.

\begin{table*}[t]
\centering
\small

\setlength{\tabcolsep}{6pt}
\renewcommand{\arraystretch}{1.1}
\begin{tabular}{l | c}
\toprule
Combination & AUROC \\
\midrule
\textbf{Hy@5 (default, z(Vc)+z(Vn@5))} & \textbf{0.752} \\
\midrule
pure Vc & 0.689 \\
pure Vn@5 & 0.696 \\
$\alpha\!=\!0.25$ ($0.25\,z(Vc)+0.75\,z(Vn))$ & 0.750 \\
$\alpha\!=\!0.75$ ($0.75\,z(Vc)+0.25\,z(Vn))$ & 0.745 \\
max(z(Vc),\, z(Vn@5)) & 0.714 \\
min(z(Vc),\, z(Vn@5)) & 0.748 \\
Vc + Vn@5 (no z-scoring) & 0.736 \\
\bottomrule
\end{tabular}
\caption{Hybrid score combination ablation for \textsc{Hy@}5. AUROC
averaged across all 30 (model, dataset) cells. $z(\cdot)$ denotes
benchmark-wise standardisation. All variants cost $5$ generations
per query.}
\label{tab:ablation_hy}
\end{table*}

\begin{table*}[t]
\centering
\small

\setlength{\tabcolsep}{6pt}
\renewcommand{\arraystretch}{1.1}
\begin{tabular}{l | ccccc}
\toprule
 & \textsc{Vc} & \textsc{Vn@}5 & \textsc{Vn@}10 & \textsc{Hy@}5 & \textsc{Hy@}10 \\
\midrule
\textsc{Vc} & 1.000 & -- & -- & -- & -- \\
\textsc{Vn@5} & 0.242 & 1.000 & -- & -- & -- \\
\textsc{Vn@10} & 0.278 & 0.650 & 1.000 & -- & -- \\
\textsc{Hy@5} & 0.762 & 0.710 & 0.587 & 1.000 & -- \\
\textsc{Hy@10} & 0.757 & 0.552 & 0.748 & 0.892 & 1.000 \\
\midrule
$\rho$ vs.\ $y$ & 0.289 & 0.285 & 0.336 & 0.356 & 0.378 \\
\bottomrule
\end{tabular}
\caption{Pairwise Spearman rank correlations between the five
per-query confidence signals, averaged across all 30 (model, dataset)
cells.  The lower triangle is shown.  Weakly-correlated signals
($\textsc{Vc}$ vs.\ $\textsc{Vn@}p$) are precisely the ones whose
$z$-scored sum produces the strongest hybrid \textsc{Hy@}$p$;
strongly-correlated signals are different sample budgets of the same
estimator family.  The bottom row reports each method's Spearman
correlation with the binary correctness label $y$ the strength of
the underlying confidence signal.}
\label{tab:correlations}
\end{table*}

\textbf{The fusion premise: Vc and Vn@$p$ are weakly correlated.}
Table~\ref{tab:correlations} reports pairwise Spearman rank
correlations between the five per-query signals, averaged across all
30 cells. \textsc{Vc} is only weakly correlated with \textsc{Vn@}5
($\rho\!=\!0.24$) and \textsc{Vn@}10 ($\rho\!=\!0.28$) -- the two
channels carry largely independent information, which is the
empirical foundation for fusing them into \textsc{Hy@}$p$. By
contrast, same-family signals are strongly correlated
(\textsc{Vn@}5 vs.\ \textsc{Vn@}10: $\rho\!=\!0.65$;
\textsc{Hy@}5 vs.\ \textsc{Hy@}10: $\rho\!=\!0.89$), confirming that
increasing $p$ refines the ranking rather than producing a
qualitatively new signal. Finally, Spearman correlation with the
binary correctness label $y$ rises from
\textsc{Vc}/\textsc{Vn@}5 ($\sim\!0.29$) to \textsc{Hy@}10 ($0.378$),
showing the same hierarchy as AUROC and confirming that the gains
come from genuine signal combination rather than coincidence.

\textbf{Embedding and clustering overhead is $\sim$1\% of LLM
cost.} Table~\ref{tab:cost_breakdown} reports the full cost
breakdown for \textsc{POOL}-\textsc{Hy@}5 at $q\!=\!0.20$, the most
expensive operating point we report. Per query, POOL incurs
$\sim$$3.3$--$5.0$ LLM generations and $\sim$$2.2$--$3.0$ embedding
calls. Even when we use the cheapest LLM in our pool
(\texttt{gpt-4.1-nano}, the choice that maximises the embedding
fraction), embedding dollars account for only $0.7$--$2.3\%$ of
total per-query cost, averaging $\mathbf{1.18\%}$. For the larger
backbones (GPT-5-mini, Claude-Haiku-4.5, Llama-3.3-70B), this
fraction drops further because LLM unit prices rise but embedding
prices are unchanged. Clustering and PSD projection are pure CPU
operations (no API calls) and complete in $< 1$ second per split on
a single core, a negligible overhead compared to LLM-call latency
($\sim$$1$--$3$s per generation). The gen/q metric used throughout
the paper is therefore a close lower bound on total cost: adding
embedding, clustering, and similarity-search overheads multiplies
it by at most $1.02$--$1.03\times$ in the worst case.

\begin{table*}[t]
\centering
\small

\setlength{\tabcolsep}{6pt}
\renewcommand{\arraystretch}{1.1}
\resizebox{\textwidth}{!}{%
\begin{tabular}{l | rr | rr | r}
\toprule
Dataset & gen/q & emb-calls/q
& LLM \$/q & embed \$/q & emb / (LLM+emb) \\
\midrule
Biology       &  4.47 &  2.37 & $\$2.67\!\times\!10^{-4}$ & $\$0.02\!\times\!10^{-4}$ &  0.79\% \\
Law           &  3.32 &  2.97 & $\$2.49\!\times\!10^{-4}$ & $\$0.06\!\times\!10^{-4}$ &  2.27\% \\
Physics       &  3.84 &  2.79 & $\$2.35\!\times\!10^{-4}$ & $\$0.03\!\times\!10^{-4}$ &  1.19\% \\
Psychology    &  4.97 &  2.40 & $\$2.94\!\times\!10^{-4}$ & $\$0.02\!\times\!10^{-4}$ &  0.69\% \\
MATH-500      &  3.56 &  2.15 & $\$2.17\!\times\!10^{-4}$ & $\$0.02\!\times\!10^{-4}$ &  0.97\% \\
\midrule
\textbf{Average} &  &  & $\boldsymbol{\$2.52\!\times\!10^{-4}}$ & $\boldsymbol{\$0.03\!\times\!10^{-4}}$ & \textbf{1.18\%} \\
\bottomrule
\end{tabular}%
}
\caption{Full cost breakdown for \textsc{POOL}-\textsc{Hy@}5 at
$q\!=\!0.20$ (the most expensive operating point reported in the
paper), per dataset. Per-query LLM cost is $\bigl((K/N)+q\bigr)\cdot 5$
generations. Per-query embedding cost is $1$ stem embedding plus the
unique answer-text embeddings needed for the sampled generations.
Prices are 2026 OpenAI public list for the cheapest LLM
(\texttt{gpt-4.1-nano}, \$$0.10$/\$$0.40$ per 1M input/output tokens)
and \texttt{text-embedding-3-small} (\$$0.02$ per 1M tokens); the
cheaper LLM choice makes the embedding fraction look as large as
possible (conservative defense). Token counts use $\sim$$0.25$
tokens/character for question stems, an 80-token prompt overhead per
LLM call, an average $120$-token generation, and a $50$-token average
answer-embedding input.}
\label{tab:cost_breakdown}
\end{table*}

\section{Per-Query Base Estimator}
\label{sec:base-estimator}

\textbf{Verbal confidence (\textsc{Vc}).}
Elicit the model's self-reported confidence alongside its answer,
normalized to $[0,1]$:
\begin{equation}
  M_{\textsc{Vc}}(x_i) = \mathrm{VC}(x_i), \qquad g_{\textsc{Vc}} = 1.
\label{eq:vc}
\end{equation}
\textsc{Vc} is cheap and prone to confidence collapse, but reflects
the model's own commitment, which sample-based estimators cannot
recover.

\textbf{Sample-diversity confidence (\textsc{Vn@}$p$).}
Draw $p$ independent answers for $x_i$, embed each with a fixed
encoder $\psi(\cdot)$, and form the Gram matrix $S \in
\mathbb{R}^{p\times p}$ with $S_{jk}=\langle\psi(\tau_j),\psi(\tau_k)
\rangle$. After PSD projection and trace-normalisation
$\rho_i = S_+/\mathrm{Tr}(S_+)$ with eigenvalues $\{\lambda_k\}$,
\begin{equation}
  M_{\textsc{Vn@}p}(x_i) = \sum_{k=1}^{p}\lambda_k\log\lambda_k,
  \qquad g_{\textsc{Vn@}p} = p.
\label{eq:vn}
\end{equation}
This is negative von Neumann entropy, so higher values (closer to
zero) indicate spectrally concentrated samples and therefore higher
confidence.

\textbf{Hybrid confidence (\textsc{Hy@}$p$).}
\textsc{Vc} fails when the model is confidently wrong;
\textsc{Vn@}$p$ fails when paraphrases of a single wrong answer
concentrate the spectrum artificially. We fuse the two channels on a
common scale,
\begin{equation}
  M_{\textsc{Hy@}p}(x_i) =
  z\!\bigl(M_{\textsc{Vc}}(x_i)\bigr) +
  z\!\bigl(M_{\textsc{Vn@}p}(x_i)\bigr),
\label{eq:hybrid}
\end{equation}
where $z(u)=(u-\bar u)/\sigma_u$ is benchmark-wise standardisation.
For \textsc{POOL}-wrapped \textsc{Hy@}$p$, the $z$-score statistics are estimated only from evaluated medoid/rescue calls, so normalization does not require scoring unevaluated queries.
\textsc{Vc} is elicited alongside one of the $p$ samples, so
$g_{\textsc{Hy@}p} = p$. (see additional experiments in Tables \ref{tab:ablation_hy} and \ref{tab:correlations} of Appendix \ref{sec:add_res}.)

\textbf{Majority agreement (\textsc{Maj@}$p$).}
The simplest discrete sampling signal: take the fraction of the $p$
samples whose parsed answer matches the modal one
\citep{wang2023selfconsistency},
\begin{equation}
  M_{\textsc{Maj@}p}(x_i) =
  \frac{1}{p}\,\bigl|\{\,j : \tau_{i,j} = \mathrm{mode}(\tau_{i,\cdot})\,\}\bigr|,
\label{eq:maj}
\end{equation}
with $g_{\textsc{Maj@}p} = p$. Strong as a default baseline but
ignores partial agreement among paraphrased answers.

\textbf{Semantic entropy (\textsc{SE@}$p$).}
The discrete-cluster analogue of \textsc{Vn@}$p$
\citep{kuhn2023semantic}: build hard semantic equivalence classes by
union-find on $\langle\psi(\tau_{i,j}),\psi(\tau_{i,k})\rangle\!\ge\!\eta$,
then take the negative Shannon entropy over cluster sizes $\{|c|\}$,
\begin{equation}
  M_{\textsc{SE@}p}(x_i) =
  \sum_c \tfrac{|c|}{p}\log\tfrac{|c|}{p},
\label{eq:se}
\end{equation}
with $g_{\textsc{SE@}p} = p$. \textsc{Vn@}$p$ is a spectral
relaxation of the same quantity that avoids the threshold $\eta$.

\section{Baseline Details}
\label{sec:baselines}

\textbf{Category 1: end-to-end baselines.}
These produce a full per-query confidence estimate and are compared against our
complete Tier-1{+}Tier-2 pipeline. \textbf{rand+kNN} is zero-label; the other
three are \emph{supervised} methods adapted from other settings to our
black-box, per-query ranking task. All three use the same
\texttt{text-embedding-3-small} stem embeddings as \textsc{POOL} and are granted
label access transductively within each cell (a deliberately generous setup):
they are trained on labeled folds and evaluated out-of-fold (repeated stratified
5-fold cross-validation), then scored by AUROC for a like-for-like comparison.
\begin{itemize}[leftmargin=1.2em,itemsep=1pt]
  \item \textbf{rand+kNN.} Evaluate the base estimator $M$ on $K$ queries chosen
    uniformly at random ($K$ = number of \textsc{POOL} medoids, so the budget
    matches), and predict every remaining query by softmax-similarity-weighted
    $k$-NN regression ($k{=}10$, $T{=}0.1$) of the evaluated scores over stem
    embeddings.

  \item \textbf{P(IK)-LR} \citep{kadavath2022language}. The
    original P(IK) fine-tunes an ``I-know'' head with access to model internals;
    since our setting is black-box, we implement it as an external
    logistic-regression probe mapping the frozen question-stem embedding to
    correctness.

  \item \textbf{P(IK)-MLP\,(q$+$a)} \citep{kadavath2022language}.
    The stronger variant: a multi-layer-perceptron probe on the concatenated
    stem and first-answer embeddings, under the same protocol.

  \item \textbf{APRICOT} \citep{ulmer-etal-2024-calibrating}. Originally a
    calibration method trained over model generations; we adapt its
    cluster-accuracy formulation to a per-query score, cluster the stem
    embeddings ($k$-means), set each query's target to its cluster's training
    accuracy, and fit a ridge regressor from stem embedding to that target,
    read out as the confidence.

    Despite their label access, these supervised probes rank poorly
(Table~\ref{tab:e2e_baselines}): predicting correctness from the stem embedding
alone without observing the model's samples is close to predicting question
\emph{difficulty}, which correlates only weakly with whether a given model gets a
given instance right. The few hundred labeled examples available per cell
compound this (the MLP overfits and trails the logistic probe), and APRICOT's
cluster-level target captures only coarse topic difficulty. Notably these methods
calibrate well but rank weakly, the opposite of what confidence \emph{ranking}
requires so even label-supervised input probes fall well short of zero-label
\textsc{POOL}, which instead reads signal directly from the model's sampled
behaviour.
\end{itemize}

\textbf{Category 2: rescue-selection baselines.}
These share our Tier-1 soft-attention propagation and differ only in the Tier-2
rescue rule: given a budget of $\lceil qN\rceil$ extra evaluations, each ranks
the non-medoid queries by a criterion and replaces the top-ranked queries'
propagated scores with their true base score.
\begin{itemize}[leftmargin=1.2em,itemsep=1pt]
  \item \textbf{Random} : uniformly random non-medoids (averaged over seeds).
  \item \textbf{Boundary} : propagated score closest to the batch median.
  \item \textbf{Confidence tails} : highest / lowest propagated score $\mu_i$.
  \item \textbf{Disagreement (ours)} : highest weighted standard deviation
    $d_i$ of the medoid scores feeding $\mu_i$.
  \item \textbf{Oracle} : highest \emph{true} propagation error
    $|\mu_i - M(x_i)|$; an upper bound requiring labels we do not have at test.
\end{itemize}

\section{Metric Details}
\label{sec:metric_details}

We evaluate confidence estimates with two complementary metrics: the
\textbf{Area Under the Receiver Operating Characteristic curve (AUROC)} for
\emph{ranking} quality and the \textbf{Expected Calibration Error (ECE)} for
\emph{calibration} quality.

\textbf{AUROC.}
For each query $x$ an estimator produces a scalar confidence score $s(x)$, which
we evaluate against the binary correctness label $y(x)\!\in\!\{0,1\}$ of the
answer. AUROC is the area under the curve traced by the true-positive rate
against the false-positive rate as the decision threshold is swept over all
values of $s$; equivalently, it is the probability that a randomly chosen correct
answer receives a higher score than a randomly chosen incorrect one. It therefore
measures only how well the scores \emph{separate} correct from incorrect outputs,
independent of any threshold: $0.5$ is chance and $1.0$ is perfect separation.
Because AUROC is invariant to any monotone rescaling of the scores, estimators
that live on different scales (e.g.\ the $[0,1]$ verbal confidence \textsc{Vc} and
the unbounded $z$-scored \textsc{Hy@}$p$) are directly comparable, which is why we
adopt it as the primary metric throughout.

\textbf{ECE.}
AUROC ignores the absolute value of a score, but deployment uses such as
abstention and routing require the scores to behave like probabilities. The
Expected Calibration Error measures this: predictions are partitioned into $M$
equal-width confidence bins $B_1,\dots,B_M$ over $[0,1]$, and
\[
\mathrm{ECE} \;=\; \sum_{m=1}^{M} \frac{|B_m|}{N}\,
\bigl|\,\mathrm{acc}(B_m) - \mathrm{conf}(B_m)\,\bigr|,
\]
where $\mathrm{acc}(B_m)$ is the empirical accuracy and $\mathrm{conf}(B_m)$ the
mean predicted confidence in bin $B_m$, $|B_m|$ is the bin population, and $N$ the
number of queries; we use $M\!=\!15$. Since \textsc{Vn@}$p$ and \textsc{Hy@}$p$
are unbounded (they are $z$-scored, not probabilities), we first map each score to
a probability by Platt scaling (a one-dimensional logistic regression). To avoid
optimistic bias, the Platt map is fit with $5$-fold cross-validation and each
score is calibrated \emph{out-of-fold} (predicted only by folds it did not train
on). For \textsc{Vc} we additionally report the raw self-reported confidence
(divided by $100$) with no fitting, to show the uncalibrated starting point.
Lower ECE is better; $0$ denotes perfect calibration.

\section{Additional Embedding Models}

\label{sec:add_embed_models}

All main experiments use \texttt{text-embedding-3-small} for stem and answer
embeddings. To check that \textsc{POOL}'s behaviour does not hinge on this
choice, we repeat the evaluation with two open-source encoders,
\texttt{bge-large-en-v1.5} and \texttt{e5-large-v2}. The ranking and the
cost--quality trade-off are preserved across all three (Table~\ref{tab:w4_embed}).

One subtlety requires care. Because $\tau$ and $\theta$ are \emph{absolute}
cosine thresholds while cosine-similarity scales differ sharply across encoders
(mean pairwise similarity ${\approx}0.17$ for \texttt{text-embedding-3-small}
versus ${\approx}0.45$--$0.50$ for \texttt{bge}/\texttt{e5}), a single fixed
$\tau$ would induce very different compression ratios and confound embedding
quality with threshold miscalibration. We therefore recalibrate $\tau$ per
encoder to a matched mean compression $K/N$ (keeping $\theta=\tau-0.15$), so
that savings are held comparable and Table~\ref{tab:w4_embed} isolates the
effect of the embedding alone.

\begin{table*}[h]
\centering            
\small
\setlength{\tabcolsep}{6pt}
\renewcommand{\arraystretch}{1.15}
\begin{tabular}{l c ccc}
\toprule
& & \multicolumn{3}{c}{\textbf{\textsc{POOL}-\textsc{Hy@}5}} \\
\cmidrule(lr){3-5}
Embedding & \textsc{Hy@}5 (base) & $q\!=\!0$ & $q\!=\!0.10$ & $q\!=\!0.20$ \\
\midrule
\texttt{text-embedding-3-small} & 0.752 & 0.702 & 0.721 & 0.736 \\
\texttt{bge-large-en-v1.5}      & 0.753 & 0.699 & 0.717 & 0.729 \\
\texttt{e5-large-v2}            & 0.755 & 0.685 & 0.704 & 0.717 \\
\bottomrule
\end{tabular}
\caption{Average AUROC across embedding encoders, with $\tau$ recalibrated per
encoder to a matched mean compression $K/N$ ($\theta=\tau-0.15$).}
\label{tab:w4_embed}
\end{table*}

\clearpage
\section{Prompts}

\begin{promptbox}{$\mathcal{P}_{\textsc{vc}}^{\text{mcq}}$: Verbal-Confidence Prompt (MMLU-Pro, Multiple-Choice)}
\textbf{You are a careful expert answering multiple-choice questions.}

\medskip
Question: \{question\}

\smallskip
Options: \ A. \{option 1\} \ \ $\cdots$ \ \ J. \{option 10\}

\smallskip
Pick the single best option. Then express your confidence as an integer
$0$--$100$. \textbf{End with two lines exactly:}
\begin{itemize}[leftmargin=1.4em,itemsep=0pt,topsep=2pt,label={}]
  \item \texttt{Answer: \textless letter\textgreater}
  \item \texttt{Confidence: \textless 0--100\textgreater}
\end{itemize}
\end{promptbox}

\begin{promptbox}{$\mathcal{P}_{\textsc{vc}}^{\text{math}}$: Verbal-Confidence Prompt (MATH-500)}
\textbf{You are a careful mathematician. Show brief reasoning, then give the
final answer in \texttt{\textbackslash boxed\{...\}}.}

\medskip
Problem: \{question\}

\smallskip
Solve the problem and report your confidence on a $0$--$100$ scale.
\textbf{End with exactly two lines:}
\begin{itemize}[leftmargin=1.4em,itemsep=0pt,topsep=2pt,label={}]
  \item \texttt{Answer: \textbackslash boxed\{\textless final answer\textgreater\}}
  \item \texttt{Confidence: \textless 0--100\textgreater}
\end{itemize}
Be calibrated: $100$ means absolutely certain, $0$ means pure guess.
\end{promptbox}

\begin{promptbox}{$\mathcal{P}_{\textsc{vc}}^{\text{trivia}}$: Verbal-Confidence Prompt (Closed-Book TriviaQA)}
\textbf{You are a careful expert. Follow the output format exactly.}

\medskip
Answer the following trivia question from your own knowledge (no documents
are provided).

\smallskip
Question: \{question\}

\smallskip
Respond with \textbf{exactly two lines and nothing else}---no explanation,
no extra words. The answer must be the shortest possible (a name, entity,
date, or short phrase):
\begin{itemize}[leftmargin=1.4em,itemsep=0pt,topsep=2pt,label={}]
  \item \texttt{Answer: \textless short answer\textgreater}
  \item \texttt{Confidence: \textless integer 0--100\textgreater}
\end{itemize}
\end{promptbox}

\begin{promptbox}{$\mathcal{P}_{\text{para}}$: Paraphrase Generation Prompt}
\textbf{You rewrite multiple-choice question stems while strictly preserving
their meaning so that the correct option does not change.} You never alter the
options, never reveal the answer, and never add or remove information that
would change which option is correct.

\medskip
You are given an MMLU-Pro \{category\} multiple-choice question. Rewrite the
\textbf{question stem} (not the options) in two different ways.

\smallskip
\textbf{Hard requirements:}
\begin{itemize}[leftmargin=1.4em,itemsep=1pt,topsep=2pt]
  \item Each rewrite must preserve the original meaning exactly.
  \item The correct option must remain unchanged. Do NOT shift difficulty,
        add hints, or eliminate options implicitly.
  \item Do NOT mention or include the options in your rewrite.
  \item Keep the rewrite a single self-contained question.
  \item Use natural fluent English; vary sentence structure and word choice;
        avoid trivial token-level edits.
  \item Each rewrite at most ${\sim}1.5\times$ the length of the original.
\end{itemize}
Return STRICT JSON: \ \texttt{\{"rephrases": ["...", "..."]\}}
\end{promptbox}

\begin{promptbox}{$\mathcal{P}_{\text{adv}}$: Adversarial Twin Generation Prompt}
\textbf{You create adversarial twin multiple-choice questions for a robustness
study on confidence estimation.} You are precise and output only valid JSON.

\medskip
\textbf{Original question} (\{category\}): \{stem\} \\
\textbf{Options:} \{options\} \\
\textbf{Correct answer:} \{gold\_letter\}. \{gold\_text\}

\smallskip
\textbf{Task.} Produce an adversarial twin such that:
\begin{enumerate}[leftmargin=1.6em,itemsep=1pt,topsep=2pt]
  \item The twin stem is a \textbf{minimal} rephrase of the original---keep
        almost all wording, structure, and length; change \textbf{exactly one}
        critical entity / quantity / condition so the correct answer changes.
  \item The twin is well-formed with exactly one correct answer.
  \item The option list keeps the original correct answer verbatim as an
        \textbf{incorrect} distractor, plus the new correct answer.
  \item The change must genuinely flip the answer---not a cosmetic edit and
        not merely inserting ``not''.
\end{enumerate}
Return JSON with keys: \ \texttt{\{twin\_stem, changed\_entity, options,
gold\_letter, new\_answer\_text, original\_answer\_letter\}}
\end{promptbox}

\end{document}